\documentclass{article}

\usepackage[preprint]{neurips_2025}

\usepackage[utf8]{inputenc} % allow utf-8 input
\usepackage[T1]{fontenc}    % use 8-bit T1 fonts
\usepackage{hyperref}       % hyperlinks
\usepackage{url}            % simple URL typesetting
\usepackage{booktabs}       % professional-quality tables
\usepackage{amsfonts}       % blackboard math symbols
\usepackage{nicefrac}       % compact symbols for 1/2, etc.
\usepackage{microtype}      % microtypography
\usepackage{xcolor}         % colors
\usepackage{tcolorbox}
\usepackage{graphicx}
\usepackage{wrapfig} 
\usepackage{enumitem}
\usepackage{booktabs, tabularx}
\usepackage{multirow}
\usepackage{pifont}
\usepackage[caption=false,font=scriptsize,labelfont=sf,textfont=sf]{subfig} 
\newtheorem{definition}{Definition}

\usepackage[ruled,linesnumbered]{algorithm2e}

\title{CUFG: Curriculum Unlearning Guided by the Forgetting Gradient}

\author{
  Jiaxing Miao \\
  Department of CS\\
  Tongji University\\
  Shanghai 201804 \\
  \texttt{JiaxingMiao@tongji.edu.cn} \\
  \And
  Liang Hu \thanks{(Corresponding author: Liang Hu.)}\\
  Department of CS\\
  Tongji University\\
  Shanghai 201804 \\
  \texttt{lianghu@tongji.edu.cn} \\
  \And
  Qi Zhang \\
  Department of CS\\
  Tongji University\\
  Shanghai 201804 \\
  \texttt{zhangqi\_cs@tongji.edu.cn} \\
  \And
  Lai Zhong Yuan \\
  Ballsnow Technology\\
  Shanghai 201804 \\
  \texttt{zhongyuan.lai@ballsnow.com} \\
  \And
  Usman Naseem \\
  the School of Computing\\
  Macquarie University\\
  Sydney, NSW 2109 \\
  \texttt{usman.naseem@mq.edu.au} \\
}

\begin{document}

\maketitle

\begin{abstract}
As privacy and security take center stage in AI, \textit{machine unlearning}—the ability to erase specific knowledge from models—has garnered increasing attention. However, existing methods overly prioritize efficiency and aggressive forgetting, which introduces notable limitations. In particular, radical interventions like gradient ascent, influence functions, and random label noise can destabilize model weights, leading to collapse and reduced reliability. To address this, we propose \textbf{CUFG} (\textbf{C}urriculum \textbf{U}nlearning via \textbf{F}orgetting \textbf{G}radients), a novel framework that enhances the stability of approximate unlearning through innovations in both forgetting mechanisms and data scheduling strategies. Specifically, CUFG integrates a new gradient corrector guided by forgetting gradients for fine-tuning-based unlearning and a curriculum unlearning paradigm that progressively forgets from easy to hard. These innovations narrow the gap with the gold-standard \textit{Retrain} method by enabling more stable and progressive unlearning, thereby improving both effectiveness and reliability. Furthermore, we believe that the concept of curriculum unlearning has substantial research potential and offers forward-looking insights for the development of the MU field. Extensive experiments across various forgetting scenarios validate the rationale and effectiveness of our approach and CUFG. Codes are available at \url{https://anonymous.4open.science/r/CUFG-6375}. 
%Notably, CUFG outperforms the state-of-the-art baselines in terms of unlearning stability and average performance gaps with Retrain. 
%随着人工智能领域对隐私和安全的日益重视，能够有效移除特定知识的机器学习方法已获得广泛关注和信任。然而，现有方法过于注重效率和激进的遗忘策略，存在显著局限性。对模型权重的激进干预往往会导致模型崩溃，从而降低稳定性和可靠性。为了解决这个问题，我们提出了 CUFG（遗忘梯度引导的课程遗忘）。具体而言，我们提出了新的遗忘机制梯度校正器和新的课程遗忘范式。这些创新实现了由易到难的稳定、渐进式遗忘，显著提升了机器学习的有效性、合理性和稳定性，缩小了近似遗忘与黄金标准 Retrain 之间的性能差距。此外，我们相信课程遗忘的概念具有巨大的研究潜力，并为 MU 领域的发展提供了前瞻性的见解。在各种遗忘场景下的大量实验验证了我们的方法和 CUFG 的合理性和有效性。值得注意的是，CUFG 在取消学习稳定性和与 Retrain 的平均性能差距方面优于现有技术基线。
\end{abstract}

\section{Introduction} \label{sec:intro}
% 近来，数据隐私、模型安全和知识可控性等问题引起广泛关注。该背景下，尊重数据“被遗忘权”的机器遗忘被催生。它旨在帮助训练后的模型有效消除特定数据点或知识的影响。当前的MU研究大致分为两类： Exact Unlearning 和 Approximate Unlearning。前者致力于研究可证明错误保证的反学习过程。代表性地，直接删除数据并retrain model的scratch方法为反学习提供了黄金标准。不幸的是，高昂的计算成本使这种方法变得不切实际，迫使研究人员走向近似的忘记学习。
Recently, concerns about data privacy, model security, and knowledge controllability have garnered widespread attention \cite{wei2025trustworthy, kande2024security}. Against this backdrop, the concept of machine unlearning (MU) has emerged, which upholds the 'right to be forgotten' for data \cite{li2025machine, liu2025rethinking}. Its objective is to enable trained models to effectively scrub the influence of specific data points or knowledge \cite{cao2015towards, yang2025erase}. Current research on MU can be broadly categorized into two approaches: exact unlearning and approximate unlearning \cite{wang2024machine, qu2024learn}. The former focuses on developing unlearning processes with provable error guarantees. Notably, the scratch method, which directly removes data and retrains the model from scratch, serves as the gold standard for unlearning. Unfortunately, its high computational cost makes this approach impractical, driving researchers toward approximate unlearning\cite{zhao2024makes, huang2024learning}.

% 近似遗忘的快速、实用性使其备受青睐。但是，其目前所采用的一些处理策略呈现出了较为显著的局限性，具体表现为对模型权重的干预过于直接和激进。如图1(b)所示,常见的干预方式有：通过Fisher信息矩阵硬性化正则调整、或者采取冻结及裁剪部分权重参数等措施后，人为引入错误标签以破坏模型原有映射关系。这些做法虽意在促进模型的遗忘过程，却往往对原始模型造成冲击或引入噪声。这导致遗忘学习的稳定性和可靠性大幅降低，引发包括但不限于模型崩溃、梯度爆炸以及梯度消失等严峻问题。

The rapidity and practicality of approximate unlearning have made it highly favored \cite{foster2024fast, yao2024machine}. However, the current processing strategies employed exhibit notable limitations, primarily due to overly direct and aggressive interventions in model weights. As illustrated in Fig.1 (b), common intervention methods include imposing rigid regularization adjustments via the Fisher information matrix \cite{golatkar2020eternal, koh2017understanding} or adopting measures such as freezing and pruning certain weight parameters \cite{jia2023model, fan2024salun}, followed by artificially introducing incorrect labels to disrupt the model's original mapping relationships \cite{golatkar2020eternal}. Although these approaches aim to facilitate the forgetting process, they often negatively impact the original model or introduce noise. This leads to a significant decline in the stability and reliability of unlearning, causing severe issues such as model collapse, gradient explosion, and gradient vanishing, among others \cite{graves2021amnesiac, thudi2022unrolling}. To address these limitations, this paper aims to tackle the following question:
\begin{tcolorbox}[before skip=3pt, after skip=2.5pt]
\emph{\textbf{(Q)} Is there an effective approximate unlearning process that is non-aggressive while simultaneously maintaining model stability and reliability?}
\end{tcolorbox}

%针对上述局限性，本文致力于解决以下问题：是否存在一种非激进式且兼顾模型稳定性与可靠性的有效近似遗忘进程？
%对于（Q），模型在遗忘进程中稳定性的考量本质上取决于遗忘机制的设计及遗忘数据本身对模型的影响。在遗忘机制上，如图1(a)和(b)所示，遗忘效率较低的黄金标准retrain是确保模型稳定收敛到保留数据局部最优解的最佳参考方案。图1(b)中的诸类高效遗忘方法因激进的权重干预，导致难以保证模型稳定；Finetune虽能保持稳定性，但仅能微弱地过拟合迁移，遗忘效果难以保证。那么，是否能从优化角度设计类似 retrain 的微调方法，实现高效且稳定的遗忘？在遗忘数据方面，模型对不同遗忘数据的自信程度不一，导致遗忘难度差异显著。实际应用中，直接输入所有遗忘指令极易冲击模型的稳定表示能力。换句话说，遗忘过程中需兼顾遗忘内容与时机。若能智能序列化处理模型对不同遗忘指令的自信度差异，减缓对原有性能的冲击，必定有助于构建合理且稳定的遗忘模型。
\setlength{\intextsep}{2pt}
\begin{wrapfigure}{r}{0.65\textwidth} \label{fig:1}% {r} 表示右对齐，宽度40%文本宽度
    \centering
    \includegraphics[width=\linewidth]{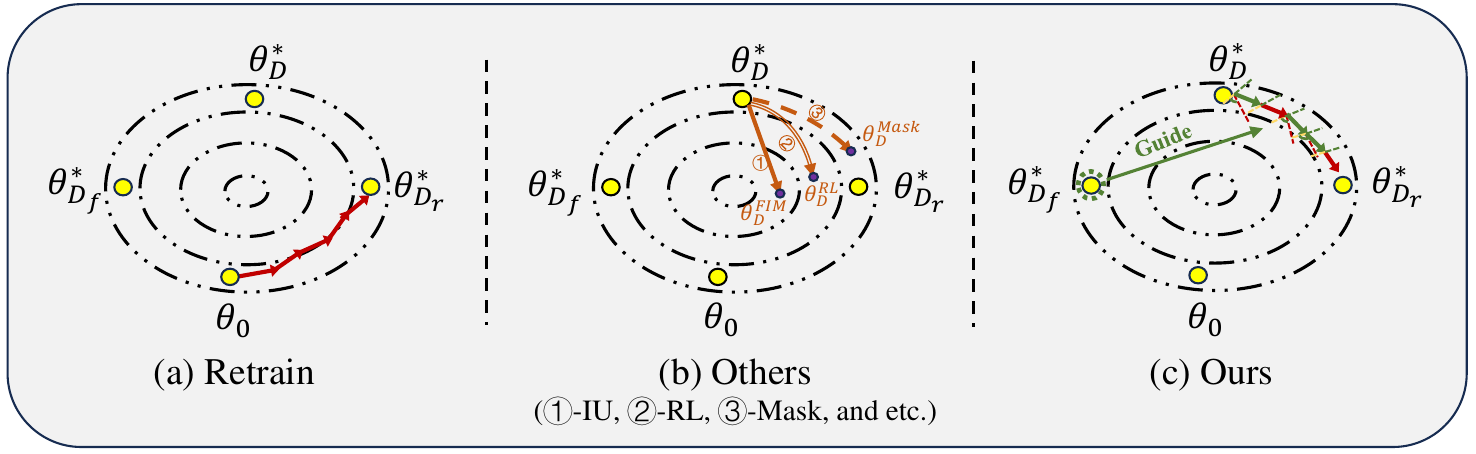}
    \caption{Different unlearning processes from an optimization perspective are shown. \textbf{Figure \ref{fig:1}(a)} shows the gold-standard (Retrain) for finding the local optimum, while \textbf{Figure \ref{fig:1}(b)} illustrates popular approximate unlearning methods, such as influence function-based IU, randomly labeled RL, and Mask-based weight sparsification or freezing methods. \textbf{Figure \ref{fig:1}(c)} is ours efficient and stable unlearning mechanism guided by forgetting gradients. %从优化的视角揭示不同方法的反学习过程。图（a）描述的是黄金标准即重新训练寻找局部最优解的过程。图（b）中展示了当前热门的近似遗忘方法的优化模式，如基于影响函数的IU，基于随机标签的RL，以及基于权重稀疏或冻结的Mask类方法。图（c）展示了本文所提方法的动机。本文提出一种受遗忘梯度引导的高效稳定的新型遗忘微调方法。其稳定性方面有类似于Retrain的功效。
    }
\end{wrapfigure}
In \textbf{(Q)}, model stability during unlearning fundamentally depends on the forgetting mechanism design and the impact of the forgotten data. Regarding the forgetting mechanism, as illustrated in Figures \ref{fig:1}(a) and \ref{fig:1}(b), the gold-standard (Retrain), despite its low forgetting efficiency, serves as the optimal reference for ensuring the model’s stable convergence to the local optimum of the retained data. In contrast, various high-efficiency unlearning methods shown in Figure \ref{fig:1}(b) often involve aggressive weight interventions, making it difficult to guarantee model stability \cite{golatkar2020eternal, koh2017understanding, jia2023model, fan2024salun}. %While fine-tuning can maintain stability, it only induces weak overfitting adaptation, leading to suboptimal forgetting performance. 
As shown in Figure \ref{fig:1}(c), it is imperative to design an efficient approximate unlearning method with stability akin to retraining. In terms of the forgotten data, the model exhibits varying degrees of confidence in different forgetting samples, resulting in significant differences in forgetting difficulty. In practice, directly inputting all forgetting instructions at once can severely disrupt the model’s stable representation ability. In other words, the unlearning process must consider both the content and timing of forgetting. Intelligent serialization of forgetting instructions based on confidence levels could %help preserve original performance and 
support stable unlearning.

% 为解决(Q)，本文针对前文所述的两个关键方面给出回应。一种受遗忘梯度指引地课程式反学习CUFG被开创性地提出。首先，被现有方法忽略的遗忘数据的推理梯度被重视。本文提出了调节Finetune优化过程的梯度修正器。它指引模型逐步趋向于保留数据上的局部最优解，而拒绝被遗忘数据影响。该范式的遗忘方案保证了模型的稳定性与遗忘效率。其次，汲取课程学习的灵感，课程反学习的概念被我们原创性地提出。智能序列化遗忘数据提供了灵感。。模型对于遗忘数据的置信程度帮助我们将遗忘数据序列化划分为不同遗忘难度的反学习课程。先易后难的过程减缓了反学习对模型稳定性的冲击。
To address \textbf{(Q)}, this paper responds to the two key aspects discussed earlier. A novel forgetting paradigm, \textbf{C}urriculum \textbf{U}nlearning guided by the \textbf{F}orgetting \textbf{G}radient (\emph{\textbf{CUFG}}), is proposed. First, the inference gradients of forgotten data are emphasized, a neglected aspect in existing methods. It guides the model toward the local optimum of retained data while resisting the influence of forgotten data. This paradigm ensures both model stability and unlearning efficiency. Second, inspired by curriculum learning \cite{wang2021survey, liu2024let, wang2024efficienttrain++}, we introduce the concept of \textbf{Curriculum Unlearning} for the first time. By progressing from easier to harder cases, this approach mitigates the impact of unlearning on model stability. \textbf{Our contributions} are summarized as follows.
\begin{itemize}[label={\normalsize$\bullet$},leftmargin=10pt,topsep=0pt,itemsep=5pt,parsep=0pt]
\item We provide a novel comprehensive perspective on the stability of approximate unlearning models in efficient unlearning by analyzing the design of forgetting mechanisms and forgotten data.
\item We propose a new fine-tuning unlearning method based on a gradient corrector (GC), inspired by the gold-standard, which leverages forgotten data inference gradients to efficiently and stably converge to the local optimum of retained data while mitigating the influence of forgotten data.
\item We develop the novel "easy-to-hard" curriculum unlearning paradigm and propose CUFG with a gradient corrector. Forgetting instructions are adaptively graded into different levels based on the model's confidence in the forgetting targets and executed via an intelligent sequence.
\item Through extensive experiments and analysis, we demonstrate that the curriculum unlearning paradigm and GC significantly enhance model stability and efficiency for approximate unlearning. Consequently, CUFG stands out among numerous competitive forgetting methods.
\end{itemize}
% 我们原创性地从遗忘机制的设计和遗忘数据两方面对近似遗忘模型在高效反学习进程中的稳定性提供了全面的思考视角。
% 我们受黄金标准Retrain的启发，通过梯度修正器改进了基于微调的遗忘方法的反学习进程。被重视的遗忘数据推理梯度帮助模型高效、稳定地优化至保留数据下的局部最优解，同时远离遗忘数据的影响。
% 我们首次提出了课程式反学习的新型遗忘范式“先易后难”，并结合梯度修正器提出了基于遗忘梯度的课程反学习。遗忘指令被根据模型对对应遗忘目标的置信程度开发为不同等级的遗忘课程，被智能化处理序列进行遗忘。

\section{Related Work} \label{sec:related}
%机器遗忘、模型微调、课程学习
\textbf{Machine unlearning.} 
%机器遗忘（Machine Unlearning）最初由 Cao 等人提出，以缓解训练后模型的隐私泄露风险，并逐步发展为消除特定数据或知识影响的新兴研究方向。Retrain 方法被视为该领域的黄金标准，能够完全撤销目标样本的影响，但计算成本高昂。为降低精确遗忘的代价，Ginart 等人基于差分隐私提出近似遗忘概念。尽管差分隐私（DP）提供了可证明的遗忘性，其固有的噪声问题限制了其实用性。近来，研究者提出了多种更有效且高效的近似遗忘方法。例如，微调（FT）通过重新训练保留数据，但可能导致灾难性遗忘；梯度上升（GA）通过调整模型参数逆转训练过程；Fisher 信息矩阵及影响函数方法估计特定样本对模型的影响，以实现有效遗忘；BE/BS 通过调整知识边界帮助模型遗忘；随机标签方法通过构建伪标签（Fake Labels）破坏模型的映射关系；在此基础上，L1 稀疏剪枝（L1-WP）和 SalUn 通过增强模型稀疏性，进一步提高数据擦除的效果。此外，机器遗忘研究已扩展至图神经网络、联邦学习等多个领域。尽管现有方法提升了反学习效率，但在模型稳定性方面仍存在明显隐患。激进的遗忘机制可能导致模型崩溃、梯度消失或梯度爆炸，影响其可用性和泛化能力。
MU, first introduced by Cao et al. \cite{cao2015towards}, aims to mitigate the privacy risks of trained models and has evolved into a new research focused on eliminating the influence of specific data or knowledge \cite{tarun2023fast, kurmanji2023towards}. Retrain is considered the gold standard in this field, as it can completely revoke the impact of target samples. To reduce the cost of exact unlearning, Ginart et al. \cite{ginart2019making} proposed the concept of approximate unlearning based on differential privacy (DP) \cite{zhang2025privacy, zhang2025cokv}. Although DP provides provable guarantees, its inherent noise issues limit practical applicability. Recently, researchers have developed more effective and efficient approximate MU methods \cite{shi2024muse, wang2025selective}. Fine-tuning (FT) \cite{golatkar2020eternal, warnecke2021machine} retrains the retained data but may cause catastrophic forgetting. Gradient ascent (GA) \cite{graves2021amnesiac, thudi2022unrolling} reverses the training process by adjusting parameters. Methods leveraging the Fisher information matrix and influence functions estimate the impact of specific samples to facilitate effective unlearning \cite{koh2017understanding, becker2022evaluating}. Boundary expansion/shrinking (BE/BS) \cite{kurmanji2023towards} aids unlearning by adjusting decision boundaries, while random labels (RL) \cite{golatkar2020eternal} disrupt the mapping by generating fake labels. Based on RL, $L_{1}$-sparse \cite{jia2023model} and SalUn \cite{fan2024salun} enhance model sparsity to improve data scrub efficacy. Moreover, MU has expanded into areas such as graph neural networks \cite{chen2022graph, li2024towards} and federated learning \cite{chen2024federated, yazdinejad2024robust}. While existing methods improve unlearning efficiency, they pose significant challenges to model stability. Aggressive MU mechanisms may lead to model collapse, gradient vanishing, or gradient explosion, affecting usability and generalization capability.

\textbf{Gradient direction optimization.}
%梯度方向优化在训练深度学习模型中起着至关重要的作用，它通过调整更新方向来提高模型收敛的稳定性和效率。经典的动量方法被广泛用于通过结合历史更新来优化梯度方向，以平滑振荡并加速收敛。更近一步地，Nesterov 加速梯度预测未来更新以主动调整梯度方向，从而实现更有效的优化。AdamW 在优化过程中考虑角度约束梯度，通过权重衰减，使梯度更新更稳定。投影梯度下降 (PGD) [8] 将梯度更新限制在可行方向上，确保约束优化任务中的稳健性。自适应梯度方向策略（例如 AdaBelief [9]）根据梯度置信度动态修改步长，从而改善非平稳环境中的学习动态。此外，由于任务与目标的区别，还衍生出梯度正交分解等其他方法。同理，若将反学习同样看作一个优化问题，那么梯度方向优化的思想是否亦有益于遗忘优化过程的有效性与稳定性。
Gradient direction optimization is crucial in deep learning training, enhancing convergence stability and efficiency by adjusting update directions \cite{qiu2025accelerating, tang2025unleashing}. Classic momentum methods smooth updates using historical gradients, while Nesterov Accelerated Gradient (NAG) \cite{sutskever2013importance} further predicts future gradient directions to optimize convergence speed. AdamW combines angle constraints and weight decay to improve gradient update stability \cite{loshchilovdecoupled}. Projected Gradient Descent (PGD) \cite{madry2018towards} restricts gradients within feasible regions, ensuring robustness in constrained optimization tasks. Adaptive gradient direction methods, such as AdaBelief \cite{zhuang2020adabelief}, adjust learning rates based on gradient confidence, optimizing learning dynamics in non-stationary environments. Additionally, due to differences in tasks and objectives, methods like gradient orthogonal decomposition \cite{xiong2022grod} have emerged. Similarly, if we view MU as an optimization problem, the idea of gradient direction optimization may also benefit the effectiveness and stability of the unlearning process.

\textbf{Curriculum learning.} 
% 课程学习由Bengio等人受人类学习方式启发提出，旨在通过设计适当训练策略（学习课程），从简单入手，逐渐帮助模型掌握复杂任务。现有的CL方法基本都符合难度度量+训练调度的一般框架，并根据其范式不同可被分为预定义的CL和自动CL。相较于预定义的CL，更智能地自动CL发展更为迅猛，代表性地包括自定进度学习、迁移教师、强化学习教师和其他等。此外，CL通常皆为超越特定任务的训练策略，其具体表现为各类即插即用的插件。因此，在各个领域CL皆有很广泛的应用,如计算机视觉、自然语言处理等。启发性地，本文认为模型对于不同遗忘指令地处理难度并不一致。本文深入探讨了由易到难的遗忘策略的价值。
Curriculum Learning (CL), proposed by Bengio et al. \cite{bengio2009curriculum} and inspired by human learning, designs progressively challenging training strategies to help models transition from simple to complex tasks. CL methods typically follow a "difficulty measure + training schedule" framework, categorized into predefined and automatic CL \cite{wang2021survey, soviany2022curriculum}. Automatic CL, which has advanced more rapidly, includes approaches like self-paced learning \cite{jiang2015self}, transfer teacher \cite{weinshall2018curriculum}, and reinforcement learning teacher \cite{narvekar2020curriculum}. As a task-agnostic strategy, CL is often implemented as plug-and-play modules with broad applications in computer vision \cite{zhou2024curbench, madan2024cl}, natural language processing \cite{lai2024autowebglm, wang2024curriculum}, etc. Inspired by CL, this work examines models' varying difficulty in processing forgetting instructions and explores progressive unlearning to enhance performance and stability.

\section{Preliminaries and Problem Statement} \label{sec:pre}
\textbf{Machine Unlearning.} The objective of the MU task is to help the trained model $h(\theta_D^*)$ scrub the influence of specific data points or advanced knowledge. The parameter $\theta_D^*$ is obtained by solving the following optimization problem on dataset $D = \left\{ (x_i, y_i) \right\}_{i=1}^M$ via empirical risk minimization:
\begin{equation}\label{ERM}
\theta_D^* = \arg\min_{\theta} \frac{1}{|D|} \sum_{(x_i, y_i) \in D} \mathcal{L}(h(x_i; \theta), y_i),
\end{equation}
where each sample $x_i$ and its corresponding label $y_i$ form a sample pair. $h(x_i, \theta)$ is the prediction of the model for sample $x_i$ under $\theta$, and $\mathcal{L}(\cdot,\cdot)$ is the loss function. The subset of data to be scrubed is called the forgetting subset, $D_f \subseteq D$, and its complement is the retained subset, $D_r = D \setminus D_f$. 

As shown in Figure \ref{fig:1}(a), $h(\theta_{D_r}^*)$ is obtained using the Scratch strategy by retraining on $D_r$ with the initial random weight $\theta_0$. Figures \ref{fig:1}(b) and \ref{fig:1}(c) illustrate that approximate unlearning seeks an efficient optimization path from $h(\theta_D^*)$ to $h(\theta_U)$, specifically scrubing $D_f$, with the goal of approximating $h(\theta_{D_r}^*)$ as closely as possible without excessive computational cost.  In other words, the influence of $D_f$ in $h(\theta_U)$ is expected to be eliminated as cleanly as possible, while the performance of the model on $D_r$ is not destroyed. Essentially, MU can be formulated as an optimization problem similar to Eq.\ref{ERM}: 
\begin{equation}\label{MU_Q}
\theta_U = \arg\min_{\theta} \left( \frac{1}{|D_r|} \sum_{(x_i, y_i) \in D_r} \mathcal{L}(h(x_i; \theta), y_i) - \lambda \cdot \frac{1}{|D_f|} \sum_{(x_j, y_j) \in D_f} \mathcal{L}(h(x_j; \theta), y_j) \right),
\end{equation}
where $\lambda$ is a balancing hyperparameter. The first and second terms in Eq. \ref{MU_Q} respectively capture MU's different requirements for the forgotten model $h(\theta_U)$  on $D_f$ and $D_r$. Typically, one serves as the main optimization objective, while the other acts as a regularization term, as seen in the difference between fine-tuning-based and random-label-based methods. \textbf{Observing Eq.\ref{ERM} and Eq.\ref{MU_Q}, the model's response to $D_f$ and $D_r$ along the optimization path appears to be the best reference for designing an effective unlearning scheme.} Furthermore, based on different forgetting scenarios, the MU task is categorized into random data forgetting and class-wise forgetting. The former removes random subsets across categories, while the latter erases all samples of a specified category.
% 公式2中前后两项分别体现了对MU对遗忘模型thetau关于Df和Dr的要求。无论按照哪一项去scrub模型，另外一项都是正则项。
% MU的任务目标为帮助训练好的模型f(θD)针对性地擦除特定数据点或高级知识的影响。其中，模型f(θD)的参数θD是通过在数据集D上对如下优化问题以经验风险最小化进行求解所得。
%其中f(x,θ)为模型在参数θ下对于样本x的预测结果，L为损失函数。MU任务中，计划擦除其影响的数据子集被称为遗忘子集，Df⊆D。而它的补集则是保留子集，即Dr=D\Df。如fig.(a)所示，f(θDr)可基于初始随机权重f(θ0)仅在Dr上retrain的Scratch策略得到模型f(θD)。如fig.(b)和(c)所示，近似反学习企图寻找到从f(θD)到针对性地近似擦除Df的反学习模型f(θu)的高效优化路径，旨在尽可能在不过度消耗计算资源的基础上逼近f(θDr)。换句话说，θu中Df的影响被希望尽可能被消除干净，而模型在Dr上的性能不被破坏。本质上，MU可以类似eq.1地表述为优化问题可被表述为。\textbf{观察eq.1和eq.2可知，优化路径中模型对于遗忘数据和保留数据的反应似乎是提出合理、高效反学习方案的最佳参考物。}此外，根据遗忘场景的不同，遗忘任务可被分为随机数据遗忘和类遗忘。前者的遗忘子集随机提取于所有类别，而后者的遗忘指令则面向指定类别的全部样本。

\textbf{Curriculum Unlearning.} CU is a novel unlearning paradigm inspired by curriculum learning, proposing a forgetting process from easy to hard \cite{bengio2009curriculum}. Initially, CL was introduced by drawing an analogy to human learning \cite{wang2021survey, soviany2022curriculum}. Over time, the general framework of "difficulty Measurer + Training Scheduler" became mainstream. Its core principle, "starting small" or "easy to hard," has been proven beneficial in machine learning. In MU research, we found that the model's confidence in different forgetting samples reveals inconsistencies in forgetting difficulty. \textbf{Insight: the idea of progressing from easy to hard may be the key to mitigating the instability caused by unlearning.} Through in-depth exploration, this paper gives the definition of the curriculum unlearning paradigm.
\begin{definition}\label{definition}
A curriculum unlearning process is a sequence of unlearning criteria over $n$ steps: $C_u = \langle T_1, \ldots, T_i, \ldots, T_n \rangle$. Each criterion $T_i$ is a reweighting of the target unlearning data distribution $P_u(x)$:
\begin{equation}\label{CU}
T_i(x) \propto W_i(x)P_u(x) \quad \forall x \in D_f,
\end{equation}
such that the following conditions are satisfied:
\begin{itemize}[leftmargin=10pt,topsep=0pt,itemsep=5pt,parsep=0pt]
\item[1.] Difficulty Monotonicity: The difficulty of each criterion, defined by a difficulty metric function $\mathcal{M}(\cdot)$, increases monotonically: $\mathcal{M}(T_i) < \mathcal{M}(T_{i+1})$.
\item[2.] Data Completeness: The union of data covered by all criteria equals the entire forgetting dataset: $\bigcup_{i=1}^n D_f^i = D_f$, where $D_f^i$ represents the forgetting data corresponding to the criteria $T_i$.
\end{itemize}
\end{definition}
% CU是本文受CL启发提出由易到难遗忘的反学习全新范式。起初，CL通过类比人类学习进程被提出。后续发展中，“难度度量”和“调度训练”的一般框架成为主流。其核心思想“从小抓起”或“由易到难”被证明有益于机器学习。\textbf{思考：模型对于不同遗忘样本的置信程度揭示了遗忘难度的不一致性。由易到难的思想或许是解开反学习破坏模型稳定性问题的关键钥匙。}深入探索后，本文对CU给出如下定义：
%一个反学习课课程定义，满足两个条件：1. 难度要递增 2. 遗忘子集的交集要等于整体遗忘数据。
\textbf{MU Evaluation.} The performance evaluation of MU requires a comprehensive assessment across multiple dimensions. The "full-stack" MU evaluation framework proposed by Jia et al. \cite{jia2023model} has been widely adopted \cite{fan2024salun}, consisting of the following key metrics. 

$\bullet$ \textit{Unlearning Accuracy (UA)} \cite{golatkar2020eternal, graves2021amnesiac}: $UA = 1 - \mathrm{Acc}(h(D_f, \theta_U))$, measuring the effectiveness of data removal by the MU method.
$\bullet$ \textit{Remaining Accuracy (RA)}: Evaluates the model’s fidelity after unlearning by assessing its performance on the remaining dataset.
$\bullet$ \textit{Test Accuracy (TA)}: Assesses the impact of MU on model generalization by evaluating \( f(\theta_U) \) on the testset.
$\bullet$ \textit{Membership Inference Attack (MIA)} \cite{song2019privacy, yeom2018privacy}: Measures the unlearning capability by determining whether forgotten data can still be inferred as part of the training set (see Appendix \ref{APP:B3}).
$\bullet$ \textit{Runtime Efficiency (RTE)}: Evaluates the computational efficiency of the MU method, reflecting its unlearning cost.

These metrics alone cannot comprehensively assess the performance of approximate unlearning methods. Instead, the absolute difference from the gold standard of perfect unlearning (Retrain) provides a more meaningful evaluation. Additionally, an aggregate metric, \textbf{Avg.Gap}, defined as the mean of all metric differences, is introduced for overall performance assessment.

% MU的性能评估需要多方面的考量。经过众多研究者深入的探索，Jia等人提出的“全栈”MU评估被广泛采纳。具体地，遗忘准确率(UA): UA = 1-acc(df)，用于评估MU方法对于目标数据影响的擦除干净程度；剩余准确率(RA): 通过MU方法在剩余数据集上面的表现评估模型被清洗时的保真程度；测试准确率(TA): 通过MU方法所得f(θu)在测试集上的表现评估MU方法对模型泛化性的影响；成员推断攻击(MIA): 通过攻击的方式确定遗忘子集数据是否属于训练数据来判断MU方法的遗忘能力，详见附录；运行时效率（RTE）：通过MU方法计算效率的衡量去考察其遗忘的代价。在上述评估体系中，单一指标数值难以准确衡量近似反学习方法的性能，其与完美遗忘的黄金标准（Retrain）之间的绝对差值更具评价意义。此外，所有指标差异的均值(avg.gap)也作为综合评估指标而引入。

\section{Methodology of CUFG}
Our CUFG originally responds to the concern of Question \textbf{(Q)} about the stability and effectiveness of approximate unlearning from the perspectives of forgetting mechanism design and data forgetting strategy. First, we provide an overview of CUFG. Then, we reveal how to use the forgetting gradient to reasonably guide unlearning. Finally, we elaborate on the details of curriculum unlearning.
%我们的CUFG原创性地从遗忘机制的设计与数据遗忘策略两方面全面地回应了问题(Q)对于近似反学习稳定性与有效性的关注。我们首先结合本文选择遗忘机制和遗忘数据策略视角的动机对本文方法CUFG进行了总体概述。我们接下来揭示了如何利用遗忘梯度对反学习进行合理地指引。最后，CUFG中课程式反学习范式的关键细节被详细阐述。
\subsection{Framework Overview}
\begin{figure}[t]  % h 表示图片尽量放在当前位置
    \centering
    \includegraphics[width=0.97\textwidth]{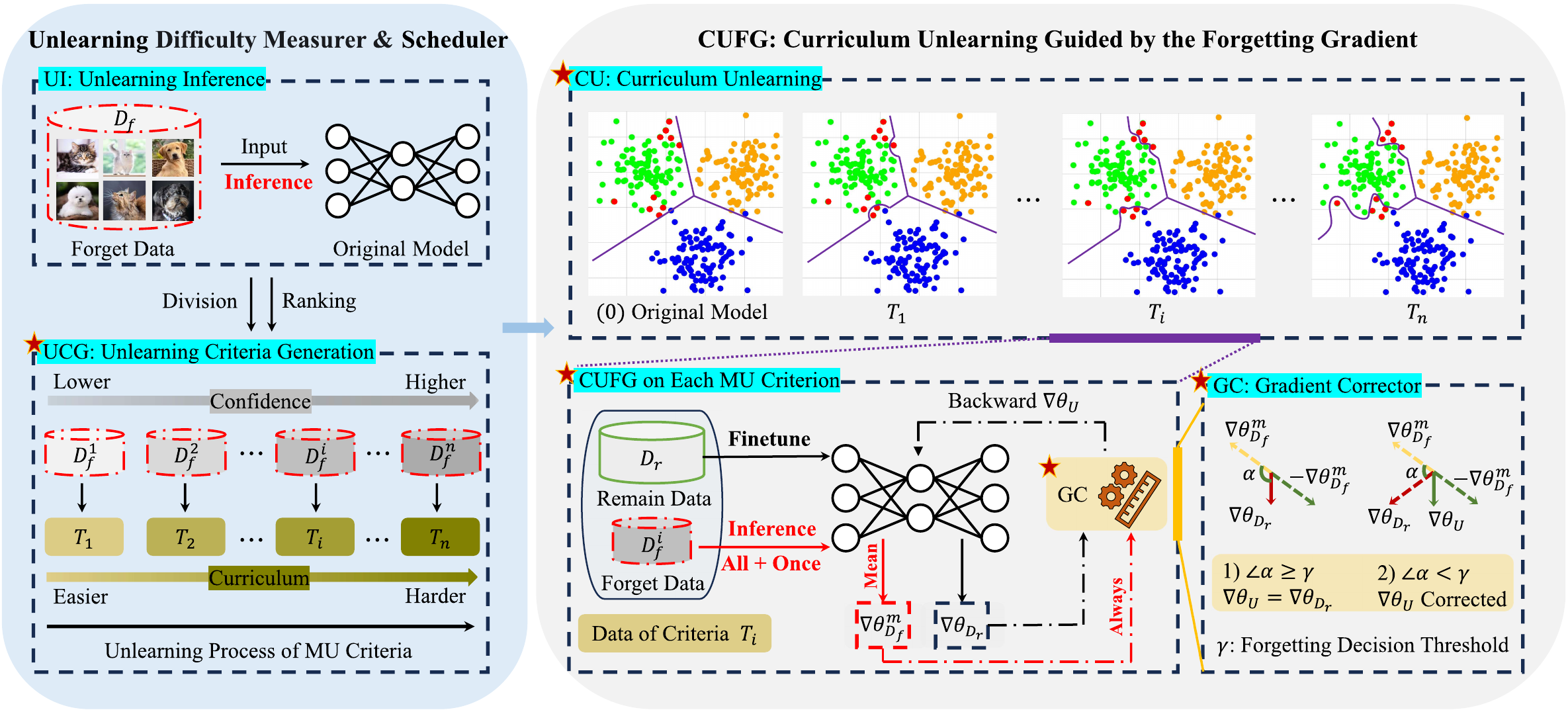}  % 替换 example-image 为你的图片文件
    \caption{Overview of the proposed CUFG. The left shows the $n$-step unlearning criteria $C_u = \langle T_1, \ldots, T_i, \ldots, T_n \rangle$, generated based on “unlearning difficulty measurer \& scheduler”. The right intuitively depicts curriculum unlearning, with the purple line highlighting the unlearning mechanism guided by the forgetting gradient. GC as the key component is emphasized by a yellow line segment.}  % 所提出的 CUFG 概览。左侧显示了基于“取消学习难度测量器 \& 调度器”生成的 $n$ 步取消学习标准 $C_u = \langle T_1, \ldots, T_i, \ldots, T_n \rangle$。右侧直观地描述了课程取消学习，紫色线突出显示了由遗忘梯度引导的取消学习机制。作为关键组件的 GC 用黄色线段突出显示。
    \label{fig:2}  % 给图片添加标签，方便引用
\end{figure}

% 图2展示了完整的CUFG架构。首先，CUFG会根据接收到的遗忘指令生成遗忘课程，即一个包含n步遗忘课程标准的序列。随后，CUFG依据该标准序列依次执行其独特的反学习机制。图中以第i个MU课程标准Ti的unlearning工作流为例，详细阐述了其遗忘机制。CUFG的设计针对性地响应了我们在遗忘机制设计及数据处理策略方面的核心考量。具体来说，图2直观地呈现了CUFG的两项创新：(1) 受遗忘梯度引导的MU方法，以及 (2) “由易到难”递进的课程式遗忘策略。相关的技术细节将在4.2节和4.3节中分别展开。值得注意的是，若仅采用(1)进行反学习，而不使用(2)，方案仍然有效，这时CUFG将退化为UFG。因此，我们在4.2节将首先探讨UFG如何有效地利用遗忘梯度引导，以实现稳定的反学习过程；而4.3节则将详细阐述课程反学习范式的具体细节，并将UFG融入该范式，从而完整呈现CUFG。
Figure \ref{fig:2} illustrates the complete CUFG architecture. CUFG first generates a forgetting curriculum from the provided forgetting instructions, forming a sequence of n-step criteria $C_u = \langle T_1, \ldots, T_i, \ldots, T_n \rangle$. Then, CUFG applies its unique  unlearning mechanism based on this sequence. The workflow of the $i$-th MU curriculum criterion, $T_i$, is shown as an example to illustrate the unlearning process. The design of CUFG addresses our core considerations regarding forgetting mechanism and data forgetting strategy, highlighted by two innovations: (1) the MU method guided by the forgetting gradient (\textbf{\emph{UFG}}), and (2) the "easier-to-harder" progressive curriculum unlearning paradigm. %The related technical details will be elaborated in Sections 4.2 and 4.3, respectively. 
Notably, using (1) alone without (2) still allows for effective unlearning, in which case CUFG degenerates into UFG. % Therefore, Section \ref{sec:UFG} will first explore how UFG effectively utilizes the forgetting gradient to achieve a stable unlearning process, while Section \ref{sec:CUFG} will provide a detailed discussion of the curriculum unlearning paradigm and integrate UFG to fully present CUFG.

\subsection{UFG: Step by Step, Fade with the Right Steps} \label{sec:UFG}
\textbf{Step by step.} The traditional backpropagation algorithm \cite{rumelhart1986learning} progressively approaches the local optimum of the deep learning optimization problem using methods such as gradient descent \cite{lecun1989backpropagation}, and its stability and effectiveness have been extensively validated. This suggests that MU might follow a "step-by-step" approach under the "Right Steps." The retraining process shown in Figure \ref{fig:1}(a) provides a reference for this. Intuitively, the process of the retrained model steadily approaching the local optimum $\theta_{D_r}^*$ from $\theta_0$ can be illustrated using the iterative gradient descent method:
%Retraining on the dataset $D_r$, the gradient direction provided by the loss function guides the model to progressively stabilize and approach the local optimum $\theta_{D_r}^*$ from the initial state $\theta_0$. Intuitively, this process is illustrated using the gradient descent method as an example:
%传统反向传播算法通过梯度下降等方法逐步逼近深度学习优化问题的局部最优解，其稳定性和有效性已得到广泛验证。这启示我们，是否MU应在“Right Steps”下呈现“Step by step”的形式。图1(a)所示的retrain给我们提供了参考。Retrain在数据集Dr上训练，损失函数反馈的梯度方向指引模型从D0出发逐步稳定地靠近局部最优解Dr*。具体如下式所述：
\begin{equation} \label{GD}
\theta^{t+1} = \theta^t - \eta \nabla \theta_{D_r}^t, 
\end{equation}
where $\nabla \theta_{D_r}^t = \nabla_\theta \left( \frac{1}{|{D_r}|} \sum_{(x_i, y_i) \in {D_r}} \mathcal{L}(h(x_i; \theta^t), y_i) \right)$. Forgetting via fine-tuning follows the same "step by step" exploration strategy as retraining, differing only in weight initialization. As anticipated, fine-tuning demonstrates advantages in terms of stability. however, its limited forgetting capability restricts its broader applicability \cite{golatkar2020eternal, warnecke2021machine}. The primary issue is that data forgetting relies solely on transfer learning, which fails to generalize due to model overfitting, and the optimization objective does not regularize the forgetting process. As a result, the gradients obtained are not the Right Steps of MU.
%Essentially, the method of forgetting via fine-tuning shares the same principles as retraining, with the only difference being the initialization of the weights. This approach follows a "step by step" exploration strategy. As anticipated, fine-tuning demonstrates advantages in terms of stability; however, its limited forgetting capability restricts its broader applicability. The primary issue lies in the fact that the consideration of data forgetting relies entirely on transfer learning, which fails to generalize the forgetting of data due to the overfitting ability of the model. Moreover, the optimization objective does not regularize the forgetting process. As a result, the feedback gradients obtained are not aligned with the Right Steps of MU.
%本质上，通过微调（finetune）进行遗忘的方法与重训（Retrain）共享上述，仅起始权重不同。其符合“Step by step”探索方式。正如我们预期的那样，微调在稳定性方面确实展现了优势，但其遗忘能力的不足限制了其推广应用。其主要问题在于对数据遗忘的思考完全依托于迁移学习是模型的过拟合能力带来的遗忘数据上泛化能力的缺失。其优化目标并未对遗忘目标进行正则化。此外，微调的优化目标并未对遗忘过程进行正则化，所得到的反馈梯度并非是针对MU的Right Steps。

\textbf{Fade with the Right Steps.} The stable performance of gradient adjustment strategies in deep learning provides insights into determining the Right Steps for the MU task. We aim to improve the gradient direction of MU fine-tuning to achieve the outcome depicted in Figure \ref{fig:1}(C). Using the intuitive explanation from Figure \ref{fig:1}, we aim for the weight distribution to closely approximate $\theta_{D_r}^*$ while staying as far as possible from $\theta_{D_f}^*$, where $\theta_{D_f}^*$ is the weight obtained from training from scratch on $D_f$. To address the limitations of MU fine-tuning in forgetting performance, we introduce a novel concept of forgetting gradients and construct a corresponding gradient corrector, leveraging regularization to help determine the Right Steps for MU.

\textbf{Forgetting Gradient.} During fine-tuning based on the retained data \( D_r \), the gradient feedback \( \nabla \theta_{D_r} \) can be directly obtained. For the forgotten data \( x_j \), the model can compute its loss and gradient response using its current weight \( \theta^t \) through inference:
%在基于保留数据 \( D_r \) 进行微调时，梯度反馈 \( \Delta D_r \) 可直接获取。而对于遗忘数据 \( x_i \)，模型可以根据当前权重\theta^t，利用推理能力计算其损失及梯度反应。
\begin{equation} \label{forgotGD}
\nabla \theta_{(x_j,y_j)}^t = \nabla_\theta \left( \mathcal{L}(h(x_j; \theta^t), y_j) \right).
\end{equation}
However, the model's gradient response to the forgotten data is inconsistent. Compared to analyzing each forgotten data point individually, the average gradient over the forgotten dataset better reflects how to quickly intensify the impact of forgotten data on the current model, specifically the gradient direction that most rapidly approaches the distribution $\theta_{D_f}^*$. Specifically, we use the "all+once" mode, where all forgotten data are input for inference at once, and the average gradient is obtained using the mean function: 
%然而，模型对于遗忘数据的梯度反应并不一致。相较于逐个分析遗忘数据，遗忘数据集上平均响应梯度更能反映出如何迅速加深遗忘数据对当前模型的影响。即最快向分布thetadf*靠近的梯度方向。具体地，我们通过all+once模式，一次性输入所有遗忘数据进行推理，并通过mean函数取得平均响应梯度。
\begin{equation} \label{meangd}
\nabla \theta_{D_f}^{m,t} = \frac{1}{|{D_f}|} \sum_{(x_j, y_j) \in {D_f}} \nabla \theta_{(x_j,y_j)}^t.
\end{equation}
To minimize the impact of forgotten data, we should aim to stay as far as possible from \( \theta_{D_f}^* \) during the optimization process. Therefore, the \textbf{forgetting gradient} \( - \nabla \theta_{D_f}^{m,t} \) is determined by selecting the opposite direction of \( \nabla \theta_{D_f}^{m,t} \). It is important to emphasize that both the forgetting gradient must be updated with $\theta^t$ in real-time during the fine-tuning process.
%若是希望尽可能地擦除遗忘数据的影响，在优化过程中远离$\theta_{D_f}^*$。我们常规地选取$\nabla \theta_{D_f}^{m,t}$的反方向作为遗忘梯度。值得注意的是，遗忘梯度和保留梯度在微调过程中都是实时更新的。

\textbf{Gradient Corrector.} Based on the above, we propose a gradient corrector (GC) guided by the forgetting gradient to identify the Right Steps for MU. Specifically, during real-time fine-tuning on data \( D_r \), the gradient may occasionally exhibit a strong component towards the distribution \( \theta_{D_f}^* \), which requires adjustment; otherwise, no correction is needed. The angle ($\angle \alpha$) between the average gradient \( \nabla \theta_{D_f}^m \) and the current fine-tuning gradient \( \nabla \theta_{D_r} \) determines whether gradient adjustment is required. This mechanism reflects the regularization strength of the hyperparameter \(\lambda\) in Eq.\ref{MU_Q}. 

In GC, we define the threshold as \(\gamma \in \left [ 0, \pi / 2 \right ]\). When $\angle \alpha < \gamma$, it indicates that the fine-tuning process is advancing towards \( \theta_{D_f}^* \) at an unacceptable rate without effectively mitigating the target influence. In this case, we adjust the direction of progress based on the forgetting gradient to steer away from \( \theta_{D_f}^* \). The unlearning gradient $\nabla \theta_{U}$ adjusted by GC is:
\begin{equation}\label{GC}
\ \angle \alpha < \gamma : \nabla \theta_U = \frac{1}{2}  (- \nabla \theta_{D_f}^m + \nabla \theta_{D_r}); \quad \quad \  \ \angle \alpha \geq \gamma : \nabla \theta_U = \nabla \theta_{D_r}
\end{equation}
%基于上述，我们提出了受遗忘梯度引导的梯度修正器来寻找MU的Right Steps。具体地，我们认为在保留数据Dr上的实时微调梯度，在部分时刻存在较强的朝向分布$\theta_{D_f}^*$的分量，需要被调整，反之则不需要。平均梯度\( \nabla \theta_{D_f}^{m,t} \)与当前微调梯度之间的夹角成为是否进行梯度调整的关键。这也是公式2超参数lambda对正则强度约束的体现。在GC中，我们设定这个角度阈值为gamma，当夹角越小时，越说明本次微调的权重分布在向df进行前进，我们需要结合遗忘梯度方向，指引我们远离df。具体操作如下：
As mentioned in the title of Section \ref{sec:UFG}, the adaptive gradient corrector constructs the Right Steps for MU, progressively guiding the optimization process and steadily mitigating the impact of forgotten targets. Ultimately, the weight update iteration for UFG is formulated as:
\begin{equation} \label{UFG}
\theta^{t+1} = \theta^t - \eta \nabla \theta_{U}^t. 
\end{equation}

%正如我们4.2节标题所述的那样，这种自适应的梯度修整器会引导我们一步步按照正确的优化步骤寻找到理想的权重分布，而遗忘那些该忘记的信息。最终我们会形成UFG的权重更新迭代式：。
\subsection{CUFG: Start with Easier, Scrub Gradually}\label{sec:CUFG}
\textbf{Start with Easier.} The concept of curriculum learning has been insightful: the learning process progresses from easier to harder, with gradual mastery \cite{bengio2009curriculum, wang2021survey}. Requiring the model to erase all forgetting targets at once during machine unlearning is clearly difficult and unreasonable. As shown in the classification example in Figure \ref{fig:2}, the model exhibits lower confidence in knowledge near the decision boundary, indicating uneven knowledge confidence. Directly executing all forgetting instructions could cause unpredictable impacts on the model. Therefore, a gradual forgetting strategy, “easier to harder”, helps minimize negative effects on the model.
%课程式学习对我们启发颇深：其学习过程从简单到难，逐步掌握。而遗忘过程却要求模型一次性全面清除。由图2中所示的分类实例可知，模型在决策边界附近的知识掌握是不自信的。显然，模型对知识的掌握程度不一。若直接执行全部遗忘指令，可能对模型产生不可预测的冲击。因此，采用逐步、由易到难的遗忘策略，有助于最小化对模型的负面影响。

\textbf{Curriculum Unlearning Paradigm.} Building on Definition \ref{definition} and the above analysis, we developed a tailored forgetting course generation paradigm for Curriculum Unlearning: \textbf{"unlearning difficulty measurer \& MU criteria scheduling."} The unlearning difficulty analysis are based on the trained model's confidence in the forgetting targets. The difficulty score sequence, $\mathrm{DM_{score}}$, can be derived from the model's inference of forgetting data, as shown in the following equation:
\begin{equation}\label{DM-score}
\mathrm{DM_{score}} = \mathcal{M}(h(\theta_D^*),D_f).
\end{equation}
%结合第三章给出的课程遗忘定义与上述分析，我们开发了专属Curriculum Unlearning的遗忘课程生成范式：”反学习难度度量+MU标准调度”。具体的反学习难度分析与度量机制依托于MU目标模型对于遗忘目标的置信度而构建。难度分数序列DM-score可由模型推理遗忘数据得出，即如下式所述：
The forgetting data queue,\( x_1, x_2, \dots, x_{\left | D_f \right |} \), is obtained by sorting $D_f$ in ascending order of $\mathrm{DM_{score}}$. Based on this queue, we construct a series of MU criteria $C_u = \langle T_1, \ldots, T_i, \ldots, T_n \rangle$. Specifically, we partition $D_f$ into $n$ subsets $D_f^1, D_f^2,\dots, D_f^n$ according to the forgetting difficulty.
\begin{equation}\label{Dfsub}
   D_f^i =  \mathcal{I}(\left\{x_1, x_2, \dots, x_{\left | D_f \right |} \right\}, i), \quad i = 1, 2, \dots, n,
\end{equation}
where $\mathcal{I}(\cdot,\cdot)$ is the slicing function based on the $\mathrm{DM_{score}}$ distribution. Each \( D_f^i \) is the data subset corresponding to the standard $T_i$. It is worth noting that this paper employs a mutually exclusive subset partitioning method, which may not be optimal. The study of curriculum unlearning still offers significant potential for further exploration, and as pioneers, we have only explored one possible approach.
%根据DM-score从低到高排序Df，得到遗忘数据队列xxxxxxx。基于该队列，我们构建了一系列MU标准。具体地，我们将Df按照遗忘难度依次等分为n个子集：
%注意的是，本文采用的是各子集互不相交的划分方式。但这不一定是最优的。关于cu的探讨是非常值得深入的。我们只是初步的尝试，作为一个引领者。

\textbf{CUFG Integration.} The above provides an effective strategy for data forgetting. Combined with UFG introduced in Section \ref{sec:UFG}, we integrate them into CUFG. Specifically, the forgetting instructions received by the trained model are first structured into a forgetting curriculum by CUFG. Then, CUFG sequentially processes each MU criterion through the UFG mechanism, gradually scrubing the impact of the forgotten data on the trained model. As shown in Figure \ref{fig:2}, the CUFG workflow emphasizes both forgetting ability and model stability. Overall, this approach offers a comprehensive and effective solution to Question (Q) from both the forgetting mechanism design and data forgetting strategy perspectives. The complete algorithm pseudocode for CUFG is provided in \textbf{Appendix \ref{APP:A}}.
%上述课程遗忘的范式为我们提供了很好的数据遗忘策略。配合我们在4.2中提出的遗忘机制UFG，我们整合了他们，提出了CUFG。具体地，已训练的模型接收到的遗忘指令首先被CUFG制作为遗忘课程。而后，CUFG通过UFG遗忘机制，依次处理各个遗忘标准。最终，遗忘数据对已训练模型的影响被逐步清除干净。如图2所示，CUFG的整个工作流都突出了MU任务对遗忘能力和模型稳定性保护两方面的保障。总的来说，从遗忘机制的设计以及数据遗忘策略两方面出发为问题Q提供了一个很全面的优秀方案。CUFG完整的算法伪代码详见附录。

\section{Experiments} \label{sec:EXP}
\subsection{Experiment Setups}
%数据集和模型、基准与衡量指标、遗忘设定与实验设置
\textbf{Data, Models and Unlearning Setups.} This paper evaluates the performance of the CUFG model in two typical forgetting scenarios in image classification \cite{liu2025comprehensive}: random data forgetting and class-wise forgetting. In the random data forgetting scenario, we tested various forgetting ratios (10\% and up to 50\%), to assess the method's effectiveness under different amounts of forgotten data. %Results were determined through 10 independent trials to ensure stability and reliability. 
To comprehensively validate its effectiveness, extensive experiments were conducted across multiple datasets (CIFAR-10, CIFAR-100, and SVHN) \cite{krizhevsky2009learning, netzer2011reading} and with different backbones (ResNet-18 and VGG-16) \cite{he2016deep, simonyan2014very}. CUFG is compared with all baselines under a unified hyperparameter setting. Its two unique hyperparameters are as follows: the angle threshold $\gamma$ in the GC module, which is determined via grid search over a specified range, and the subset number $n$ for the MU curriculum, which is empirically selected based on the distribution of data forgetting difficulty. The remaining detailed experimental setup can be found in \textbf{Appendix \ref{APP:B2}}.
% 我们主要基于图像分类任务中的两种遗忘场景：随机数据遗忘以及class-wise遗忘，对CUFG的性能进行了评估。为多方面验证其有效性，我们再多个数据集：CIFAR-10，CIFAR-100以及SVHN，与不同的backbone：ResNet-18，VGG-16上进行了大量实验。我们在本章展示了展示了较为主要的部分，其余实验详见附录。
%在随机数据遗忘场景中，随机数据的比例可以是10%或更多的50%，用于验证遗忘方法在面临不同遗忘数据量时的有效性。我们以10次随机试验确定最终的实验结果。CUFG与所有的Baselines均在公平的超参数环境下进行对比。其两个独特的超参数：GC中自适应判定的角度阈值\gamma，以及遗忘课程划分的子集数量n皆依据数据遗忘难度经验性地选取。

\textbf{Baselines and Evaluation.} In the experiments, we selected widely influential baselines in the MU field, including representative traditional methods (Retrain, FT \cite{warnecke2021machine}, GA \cite{graves2021amnesiac, thudi2022unrolling}) and state-of-the-art efficient forgetting methods (IU \cite{koh2017understanding, becker2022evaluating}, BE \cite{kurmanji2023towards}, BS \cite{kurmanji2023towards}, L1-sparse \cite{jia2023model}, and SalUn \cite{fan2024salun}). These baselines will be comprehensively compared with the proposed methods: \textbf{UFG and CUFG}. The relationship between the proposed problem Q and these methods is thoroughly analyzed in Sections \ref{sec:intro} and \ref{sec:related}. For performance evaluation of MU methods, Retrain is used as the golden standard, with a comprehensive assessment based on the metrics discussed in Section \ref{sec:pre}. Additionally, the average gap with Retrain serves as a key reference. Stability of MU methods is evaluated through model unlearning curves and sensitivity analysis of metrics. Notably, we designed some experiments specifically to validate the innovative motivation behind CUFG.
%，实验中，我们选取了当前MU领域内影响力广泛的一些基线方法，其中包括代表性的传统方法FT、GA，以及star of the art的新高效机器遗忘方法IU，BE、BS，L1-sparse和SalUn。问题Q的提出与这些方法的关联，我们在Introduction以及related work中进行了有见解的分析。在MU方法的性能评估方面，我们以Retrain为黄金标准，主要使用了第三节中提到的各个指标进行综合评估。此外，与Retrain之间的平均差距，被作为一个很重要的参考标准。对于MU方法的稳定性来讲，我们通过各指标敏感性分析、模型反学习曲线等方面进行评估。值得注意的是，本节中，CUFG的创新动机也得到了验证。

\subsection{Experiment Results}
\begin{table}[ht]
\caption{\small Performance overview of various MU methods (including the proposed UFG and CUFG) in image classification tasks on the CIFAR-10 dataset. The table presents four different forgetting scenarios: Random Data Forgetting (10\%) on ResNet-18, Random Data Forgetting (50\%) on ResNet-18, Random Data Forgetting (10\%) on VGG-16, and Class-wise Forgetting on ResNet-18. %Evaluation metrics follow the approach described in Section \ref{sec:pre}. 
The performance gap with Retrain is represented by the \textcolor{blue}{blue} values, with a smaller gap indicating better performance of the MU method. \textbf{To highlight the  different forgetting scenarios, only selected results on CIFAR-10.} Additional experimental results in \textbf{Appendix \ref{APP:C}}.}
\label{performance}
%表1：CIFAR-10数据集上各MU方法在图像分类任务中的性能总结。表中展示了4种不同的遗忘场景：Random Data Forgetting (10\%) on ResNet-18、Random Data Forgetting (50\%) on ResNet-18、Random Data Forgetting (10\%) on VGG-16、Class-wise Forgetting on ResNet-18。评估指标以第三章陈述的方案为准。与Retrain的性能差距由蓝色数值表示。请注意，MU方法的性能越好，与Retrain的性能差距就越小。其他更多实验结果被展示于附录。
\scalebox{0.48}{
\begin{tabular*}{2.05\textwidth}{@{}p{2cm}<{\centering}|p{2cm}<{\centering}p{2cm}<{\centering}p{2cm}<{\centering}p{2cm}<{\centering}p{2cm}<{\centering}|p{2cm}<{\centering}|p{2cm}<{\centering}p{2cm}<{\centering}p{2cm}<{\centering}p{2cm}<{\centering}p{2cm}<{\centering}@{}}
\toprule [2pt]
\multirow{2}{*}{\textbf{Methods}} & \multicolumn{5}{c|}{\textbf{Random Data Forgetting (10\%) on ResNet-18}}                                                                                               & \multirow{2}{*}{\textbf{Methods}} & \multicolumn{5}{c}{\textbf{Random Data Forgetting   (50\%) on ResNet-18}}                                                                                                 \\ [2pt]
                         & \multicolumn{1}{c|}{\textbf{UA}} & \multicolumn{1}{c|}{\textbf{RA}} & \multicolumn{1}{c|}{\textbf{TA}} & \multicolumn{1}{c|}{\textbf{MIA}}   & \textbf{Avg.Gap} &                          & \multicolumn{1}{c|}{\textbf{UA}} & \multicolumn{1}{c|}{\textbf{RA}} & \multicolumn{1}{c|}{\textbf{TA}} & \multicolumn{1}{c|}{\textbf{MIA}}    & \textbf{Avg.Gap} \\ [2pt] \midrule [1pt]
Retrain                  & 8.04 (\textcolor{blue}{0.00})                            & 100.00 (\textcolor{blue}{0.00})                          & 91.39 (\textcolor{blue}{0.00})                           & \multicolumn{1}{p{2cm}<{\centering}|}{16.60 (\textcolor{blue}{0.00})}          & \textcolor{blue}{0.00}             & Retrain                  & 11.91 (\textcolor{blue}{0.00})                            & 100.00 (\textcolor{blue}{0.00})                          & 87.81 (\textcolor{blue}{0.00})                           & \multicolumn{1}{p{2cm}<{\centering}|}{22.71 (\textcolor{blue}{0.00})}           & \textcolor{blue}{0.00}             \\ [2pt]
FT                 & 1.96 (\textcolor{blue}{6.08})                            & 99.44 (\textcolor{blue}{0.56})                           & 92.93 (\textcolor{blue}{1.54})                           & \multicolumn{1}{c|}{5.20 (\textcolor{blue}{11.40})}           & \textcolor{blue}{4.90}             & FT                 & 0.98 (\textcolor{blue}{10.93})                            & 99.84 (\textcolor{blue}{0.16})                           & 93.60 (\textcolor{blue}{5.79})                           & \multicolumn{1}{c|}{3.78 (\textcolor{blue}{18.93})}            & \textcolor{blue}{8.95}             \\ [2pt]
GA               & 1.00 (\textcolor{blue}{7.04})                            & 99.34 (\textcolor{blue}{0.66})                           & 93.68 (\textcolor{blue}{2.29})                           & \multicolumn{1}{c|}{2.09 (\textcolor{blue}{14.51})}           & \textcolor{blue}{6.13}             & GA               & 0.59 (\textcolor{blue}{11.32})                            & 99.43 (\textcolor{blue}{0.57})                           & 94.54 (\textcolor{blue}{6.73})                           & \multicolumn{1}{c|}{12.13 (\textcolor{blue}{10.58})}           & \textcolor{blue}{7.30}             \\ [2pt]
IU              & 1.06 (\textcolor{blue}{6.98})                            & 99.65 (\textcolor{blue}{0.35})                           & 93.84 (\textcolor{blue}{2.45})                           & \multicolumn{1}{c|}{2.46 (\textcolor{blue}{14.14})}           & \textcolor{blue}{5.98}             & IU              & 3.77 (\textcolor{blue}{8.14})                            & 97.48 (\textcolor{blue}{2.52})                           & 90.74 (\textcolor{blue}{2.93})                           & \multicolumn{1}{c|}{5.52 (\textcolor{blue}{17.19})}            & \textcolor{blue}{7.70}             \\ [2pt]
BE               & 2.32 (\textcolor{blue}{5.72})                            & 99.32 (\textcolor{blue}{0.68})                           & 93.46 (\textcolor{blue}{2.07})                           & \multicolumn{1}{c|}{8.62 (\textcolor{blue}{7.98})}           & \textcolor{blue}{4.11}             & BE               & 3.51 (\textcolor{blue}{8.40})                           & 96.16 (\textcolor{blue}{3.84})                           & 91.55 (\textcolor{blue}{3.74})                           & \multicolumn{1}{c|}{27.93 (\textcolor{blue}{5.22})}           & \textcolor{blue}{5.30}             \\ [2pt]
BS               & 3.28 (\textcolor{blue}{4.76})                           & 99.59 (\textcolor{blue}{0.41})                           & 92.69 (\textcolor{blue}{1.30})                           & \multicolumn{1}{c|}{9.46 (\textcolor{blue}{7.14})}           & \textcolor{blue}{3.40}             & BS               & 8.78 (\textcolor{blue}{3.13})                           & 90.45 (\textcolor{blue}{9.55})                           & 83.47 (\textcolor{blue}{4.34})                           & \multicolumn{1}{c|}{35.35 (\textcolor{blue}{12.64})}           & \textcolor{blue}{7.42}             \\ [2pt]
L1-sparse                & 5.51 (\textcolor{blue}{2.53})                           & 97.01 (\textcolor{blue}{2.99})                           & 91.13 (\textcolor{blue}{0.26})                           & \multicolumn{1}{c|}{11.49 (\textcolor{blue}{5.11})}          & \textcolor{blue}{2.72}             & L1-sparse                & 1.73 (\textcolor{blue}{10.18})                           & 99.52 (\textcolor{blue}{0.48})                           & 92.69 (\textcolor{blue}{4.88})                           & \multicolumn{1}{c|}{5.41 (\textcolor{blue}{17.30})}            & \textcolor{blue}{8.21}             \\ [2pt]
SalUn                    & 3.98 (\textcolor{blue}{4.06})                           & 98.97 (\textcolor{blue}{1.03})                           & 93.08 (\textcolor{blue}{1.69})                           & \multicolumn{1}{c|}{13.51 (\textcolor{blue}{3.09})}          & \textcolor{blue}{2.47}             & SalUn                    & 5.41 (\textcolor{blue}{6.50})                           & 96.39 (\textcolor{blue}{3.61})                           & 90.89 (\textcolor{blue}{3.08})                           & \multicolumn{1}{c|}{14.16 (\textcolor{blue}{8.55})}           & \textcolor{blue}{5.44}             \\ [2pt] \midrule [1pt]
\textbf{UFG}         & \textbf{6.76 (\textcolor{blue}{1.28})}                   & \textbf{98.73 (\textcolor{blue}{1.27})}                   & \textbf{91.97 (\textcolor{blue}{0.58})}                   & \multicolumn{1}{c|}{\textbf{10.22 (\textcolor{blue}{6.38})}} & \textbf{\textcolor{blue}{2.38}}    & \textbf{UFG}         & \textbf{6.84 (\textcolor{blue}{5.07})}                   & \textbf{96.95 (\textcolor{blue}{3.05})}                   & \textbf{89.35 (\textcolor{blue}{1.54})}                   & \multicolumn{1}{c|}{\textbf{11.36 (\textcolor{blue}{11.35})}}  & \textbf{\textcolor{blue}{5.25}}    \\ [2pt]
\textbf{CUFG}     & \textbf{6.51 (\textcolor{blue}{1.53})}                   & \textbf{98.16 (\textcolor{blue}{1.84})}                   & \textbf{91.36 (\textcolor{blue}{0.03})}                   & \multicolumn{1}{c|}{\textbf{11.29 (\textcolor{blue}{5.31})}} & \textbf{\textcolor{blue}{2.18}}    & \textbf{CUFG}     & \textbf{7.24 (\textcolor{blue}{4.67})}                   & \textbf{96.43 (\textcolor{blue}{3.57})}                   & \textbf{88.92 (\textcolor{blue}{1.11})}                   & \multicolumn{1}{c|}{\textbf{11.62 (\textcolor{blue}{11.09})}}  & \textbf{\textcolor{blue}{5.11}}    \\ [2pt] \midrule [2pt]
\multirow{2}{*}{\textbf{Methods}} & \multicolumn{5}{c|}{\textbf{Random Data Forgetting (10\%) on VGG-16}}                                                                                                    & \multirow{2}{*}{\textbf{Methods}} & \multicolumn{5}{c}{\textbf{Class-wise Forgetting on ResNet-18}}                                                                                                           \\ [2pt]
                         & \multicolumn{1}{c|}{\textbf{UA}} & \multicolumn{1}{c|}{\textbf{RA}} & \multicolumn{1}{c|}{\textbf{TA}} & \multicolumn{1}{c|}{\textbf{MIA}}   & \textbf{Avg.Gap} &                          & \multicolumn{1}{c|}{\textbf{UA}} & \multicolumn{1}{c|}{\textbf{RA}} & \multicolumn{1}{c|}{\textbf{TA}} & \multicolumn{1}{c|}{\textbf{MIA}}    & \textbf{Avg.Gap} \\ [2pt] \midrule [1pt]
Retrain                  & 5.20 (\textcolor{blue}{0.00})                           & 100.00 (\textcolor{blue}{0.00})                          & 94.36 (\textcolor{blue}{0.00})                           & \multicolumn{1}{c|}{13.64 (\textcolor{blue}{0.00})}          & \textcolor{blue}{0.00}             & Retrain                  & 100.00 (\textcolor{blue}{0.00})                            & 100.00 (\textcolor{blue}{0.00})                          & 94.41 (\textcolor{blue}{0.00})                           & \multicolumn{1}{c|}{100.00 (\textcolor{blue}{0.00})}          & \textcolor{blue}{0.00}             \\ [2pt]
FT                 & 2.20 (\textcolor{blue}{3.00})                           & 99.15 (\textcolor{blue}{0.85})                           & 91.56 (\textcolor{blue}{2.80})                           & \multicolumn{1}{c|}{5.11 (\textcolor{blue}{8.53})}           & \textcolor{blue}{3.80}             & FT                 & 31.85 (\textcolor{blue}{68.15})                           & 99.82 (\textcolor{blue}{0.18})                           & 93.87 (\textcolor{blue}{0.54})                           & \multicolumn{1}{c|}{87.07 (\textcolor{blue}{12.93})}           & \textcolor{blue}{20.45}            \\ [2pt]
GA               & 0.82 (\textcolor{blue}{4.38})                           & 99.42 (\textcolor{blue}{0.58})                           & 93.20 (\textcolor{blue}{1.16})                           & \multicolumn{1}{c|}{1.11 (\textcolor{blue}{12.53})}           & \textcolor{blue}{4.66}             & GA               & 97.69 (\textcolor{blue}{2.31})                             & 95.13 (\textcolor{blue}{4.87})                           & 88.82 (\textcolor{blue}{5.59})                           & \multicolumn{1}{c|}{98.36 (\textcolor{blue}{1.64})}           & \textcolor{blue}{3.60}             \\ [2pt]
IU             & 1.98 (\textcolor{blue}{3.22})                           & 98.21 (\textcolor{blue}{1.79})                           & 90.69 (\textcolor{blue}{3.67})                           & \multicolumn{1}{c|}{3.18 (\textcolor{blue}{10.46})}           & \textcolor{blue}{4.79}             & IU              & 93.16 (\textcolor{blue}{6.84})                             & 96.84 (\textcolor{blue}{3.16})                           & 90.42 (\textcolor{blue}{3.99})                           & \multicolumn{1}{c|}{97.42 (\textcolor{blue}{2.58})}           & \textcolor{blue}{4.14}             \\ [2pt]
BE               & 0.80 (\textcolor{blue}{4.40})                           & 99.38 (\textcolor{blue}{0.62})                           & 92.47 (\textcolor{blue}{1.89})                           & \multicolumn{1}{c|}{1.83 (\textcolor{blue}{11.81})}           & \textcolor{blue}{4.68}             & BE               & 77.16 (\textcolor{blue}{22.84})                            & 98.02 (\textcolor{blue}{1.98})                           & 92.04 (\textcolor{blue}{2.37})                           & \multicolumn{1}{c|}{93.54 (\textcolor{blue}{6.46})}           & \textcolor{blue}{8.41}             \\ [2pt]
BS               & 0.86 (\textcolor{blue}{4.34})                           & 99.40 (\textcolor{blue}{0.60})                           & 92.65 (\textcolor{blue}{1.71})                           & \multicolumn{1}{c|}{1.76 (\textcolor{blue}{11.88})}           & \textcolor{blue}{4.63}             & BS               & 78.05 (\textcolor{blue}{21.95})                           & 97.87 (\textcolor{blue}{2.13})                           & 92.08 (\textcolor{blue}{2.33})                           & \multicolumn{1}{c|}{93.47 (\textcolor{blue}{6.53})}           & \textcolor{blue}{8.23}             \\ [2pt]
L1-Sparse                & 5.53 (\textcolor{blue}{0.33})                           & 97.01 (\textcolor{blue}{2.99})                           & 90.00 (\textcolor{blue}{4.36})                           & \multicolumn{1}{c|}{11.13 (\textcolor{blue}{2.51})}          & \textcolor{blue}{2.55}             & L1-Sparse                & 100.00 (\textcolor{blue}{0.00})                            & 97.01 (\textcolor{blue}{2.99})                           & 90.96 (\textcolor{blue}{3.45})                           & \multicolumn{1}{c|}{100.00 (\textcolor{blue}{0.00})}          & \textcolor{blue}{1.61}             \\ [2pt]
SalUn                    & 4.36 (\textcolor{blue}{0.84})                           & 98.11 (\textcolor{blue}{1.89})                           & 90.91 (\textcolor{blue}{3.45})                           & \multicolumn{1}{c|}{9.20 (\textcolor{blue}{4.44})}           & \textcolor{blue}{2.66}             & SalUn                    & 98.71 (\textcolor{blue}{1.29})                             & 99.65 (\textcolor{blue}{0.35})                           & 94.68 (\textcolor{blue}{0.27})                           & \multicolumn{1}{c|}{100.00 (\textcolor{blue}{0.00})}          & \textcolor{blue}{0.48}             \\ [2pt] \midrule
\textbf{UFG}         & \textbf{5.51 (\textcolor{blue}{0.31})}                   & \textbf{98.81 (\textcolor{blue}{1.19})}                   & \textbf{90.81 (\textcolor{blue}{3.55})}                   & \multicolumn{1}{c|}{\textbf{9.07 (\textcolor{blue}{4.57})}}  & \textbf{\textcolor{blue}{2.41}}    & \textbf{UFG}         & \textbf{100.00 (\textcolor{blue}{0.00})}                    & \textbf{99.68 (\textcolor{blue}{0.32})}                   & \textbf{92.87 (\textcolor{blue}{1.54})}                   & \multicolumn{1}{c|}{\textbf{100.00 (\textcolor{blue}{0.00})}} & \textbf{\textcolor{blue}{0.47}}    \\ [2pt]
\textbf{CUFG}     & \textbf{4.98 (\textcolor{blue}{0.22})}                   & \textbf{98.84 (\textcolor{blue}{1.16})}                   & \textbf{91.46 (\textcolor{blue}{2.90})}                   & \multicolumn{1}{c|}{\textbf{9.27 (\textcolor{blue}{4.37})}}  & \textbf{\textcolor{blue}{2.16}}    & \textbf{CUFG}     & \textbf{\textbf{100.00 (\textcolor{blue}{0.00})}}                    & \textbf{99.43 (\textcolor{blue}{0.57})}                   & \textbf{93.17 (\textcolor{blue}{1.24})}                   & \multicolumn{1}{c|}{\textbf{100.00 (\textcolor{blue}{0.00})}} & \textbf{\textcolor{blue}{0.45}}    \\ [2pt] \bottomrule [2pt]
\end{tabular*}}
\end{table}
% 如表1所示，我们基于评估指标UA、RA、TA、以及MIA,将本文提出的CUFG及其退化版本UFG与start of the art的baselines的MU性能进行了全方位的比较。单一的指标并不能说明方法的先进性，因为部分方法是通过牺牲RA、TA为代价进行遗忘的。与Retrain间的各指标差距及其均值Avg.Gap，刻画了与Retrain性能差距的缩小。我们展示了四个不同的遗忘场景设置，分别是(i)(ii)(iii)(iv)。值得注意的是，表1仅展示了在CIFAR-10上的部分实验。其余更多详见附录。表1中的实验结果表明：
\textbf{Performance of MU in Different Forgetting Scenarios.} As shown in Table \ref{performance}, we conduct a comprehensive comparison of the MU performance of the proposed CUFG and its degraded version UFG, against state-of-the-art baselines using the evaluation metrics UA, RA, TA, and MIA. A single metric is insufficient to assess the method's effectiveness, as some approaches trade off RA and TA for forgetting. The performance gap with Retrain, along with the average gap (Avg.Gap), reflects the reduction in performance disparity. %Four distinct forgetting scenarios are presented: Random Data Forgetting (10\%) on ResNet-18, Random Data Forgetting (50\%) on ResNet-18, Random Data Forgetting (10\%) on VGG-16, and Class-wise Forgetting on ResNet-18. 
%Note that Table \ref{performance} presents only a subset of experiments, with additional details \textcolor{red}{in the appendix}. 
\textbf{Please note that in order to highlight the stability advantages of UFG and CUFG in different forgetting scenarios, the table \ref{performance} only shows a subset of experiments. For more details, please refer to Appendix \ref{APP:C}.}
The results in Table \ref{performance} indicate:

%(I) 在所有的遗忘场景中，相较于最先见基线方法，CUFG均取得了与Retrain的平均性能差距(Avg.Gap)最小的优秀结果。其退化版本UFG的表现仅次于它。稳定的遗忘机制以及数据遗忘策略帮助CUFG很好地Unlearning(UA+MIA)，并平衡了模型在保留数据集上的性能(RA)以及泛化性能(TA)。
%(II) 从10%到50%的随机数据遗忘对比可知，部分方法的性能受到遗忘数据量变化的较大影响，如L1-sparse和BS。而我们的UFG以及CUFG均表现出稳健、令人安心的性能，尤其是在模型遗忘精度UA以及模型泛化能力TA上。
%(III)  在不同Backbone(ResNet18和VGG-16)上的实验结果在各个指标下均展现出强劲的性能，没有显示明显的漏洞。这说明，UFG和CUFG是模型无关的通用方法，可以被广泛的应用。
%(IV)  CUFG与UFG在Class-wise遗忘场景下的性能也得到了验证。UA以及MIA报告的精度解释了对某类信息的全面清除，而RA、TA，以及Avg.Gap均尽可能地向Retrain展现出的黄金标准逼近。
%(V)  注意，UFG和CUFG的比较是对我们所提出方案(受遗忘梯度指引的遗忘机制和课程反学习策略)有效性的消融验证。
\textbf{\ding{72} (I)} In all forgetting scenarios, CUFG consistently achieves the smallest average performance gap (Avg.Gap) compared to the state-of-the-art baseline methods, delivering outstanding results similar to Retrain. Its degraded version, UFG, delivers strong results, ranking just below CUFG. The stable forgetting mechanism and data forgetting strategy in CUFG effectively enhance Unlearning (UA \& MIA), while balancing performance on retained data (RA) and generalization ability (TA). \textbf{\ding{72} (II)} The comparison between 10\% and 50\% random data forgetting shows that some methods, such as L1-sparse and BS, are significantly affected by the amount of forgotten data. In contrast, both UFG and CUFG demonstrate robust and reliable performance, particularly in terms of model unlearning accuracy (UA) and generalization (TA). \textbf{\ding{72} (III)} Experiments on different backbones (ResNet-18 and VGG-16) show strong performance across all metrics, without any noticeable weaknesses. This suggests that UFG and CUFG are model-agnostic approaches, making them widely applicable. \textbf{\ding{72} (IV)} The performance of CUFG and UFG in the Class-wise forgetting scenario has also been validated. The accuracies reported for UA and MIA indicate comprehensive removal of certain class-wise information, while RA, TA, and Avg.Gap demonstrate that the methods approach the golden standard set by Retrain. \textbf{\ding{72} (V)} Notably, the comparison between UFG and CUFG serves as ablation studies to validate the effectiveness of our proposed approach, which combines a forgetting mechanism guided by the forgetting gradient  with the curriculum unlearning strategy.

%----------------------------------------------
\begin{figure}[ht]
	\setlength{\abovecaptionskip}{-0.5cm}
	\setlength{\belowcaptionskip}{0cm}
	\begin{center}
		\subfloat[Instability Analysis (Avg.Gap)]{
			\includegraphics[width=0.245\textwidth, height=2.8cm]{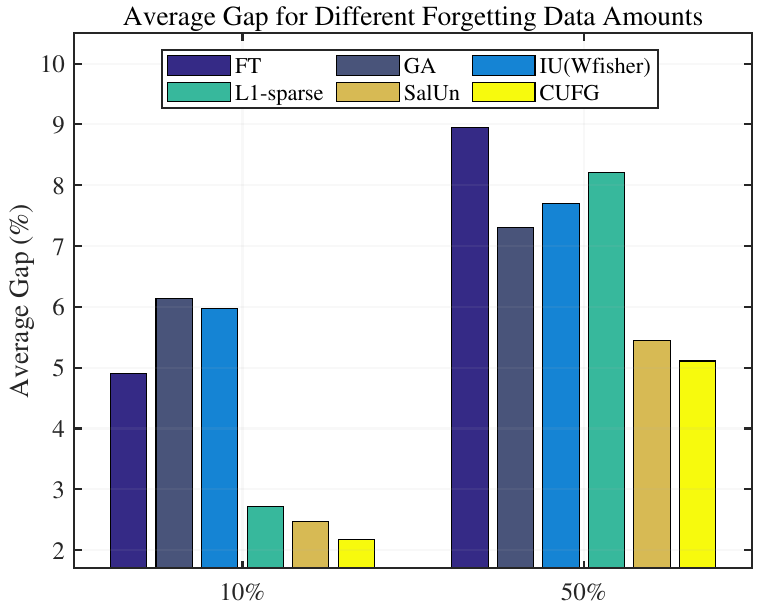}}
		\subfloat[Instability Analysis (UA)]{
			\includegraphics[width=0.245\textwidth, height=2.8cm]{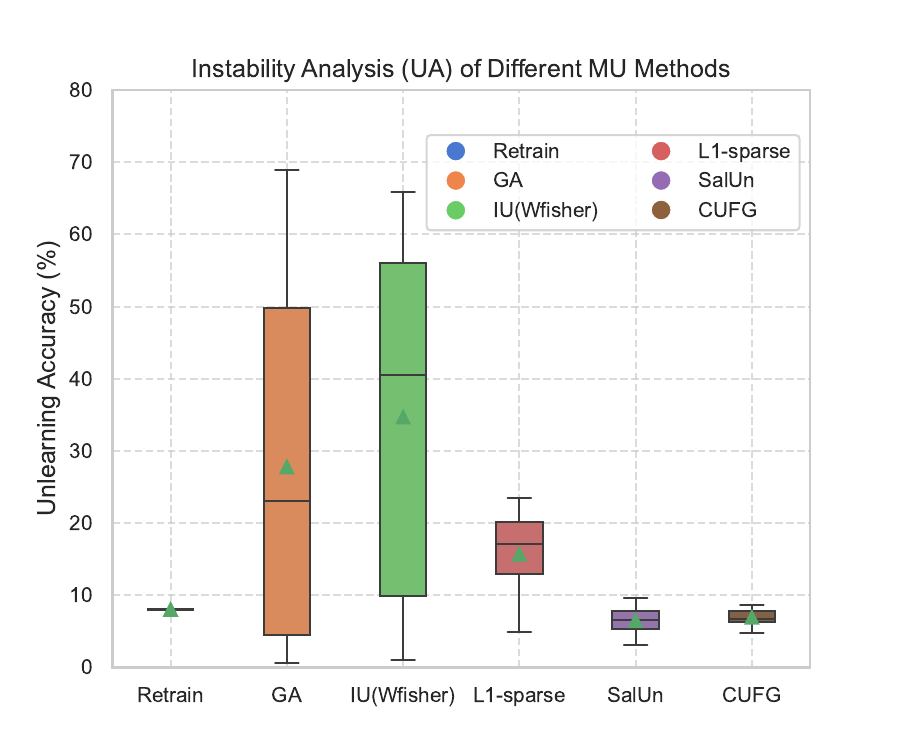}}
            \subfloat[Instability Analysis (TA)]{
			\includegraphics[width=0.245\textwidth, height=2.8cm]{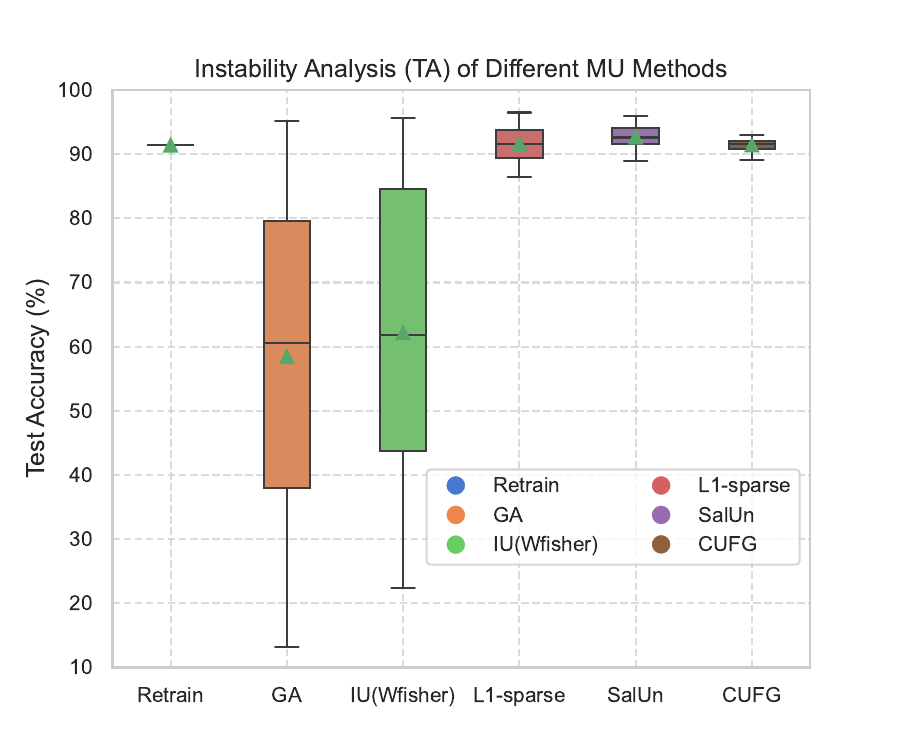}}
            \subfloat[Unlearning Curves]{
			\includegraphics[width=0.245\textwidth, height=2.8cm]{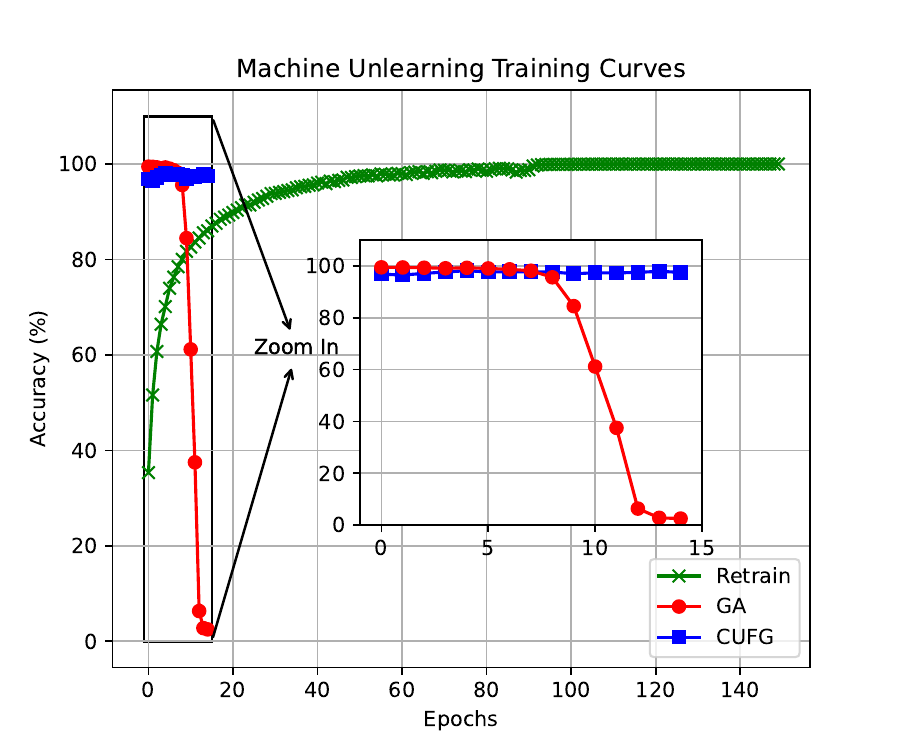}}
	\end{center}
        \vspace{0.5cm}
	\caption{\small Instability analysis of MU in different standards. \textbf{(a)} shows the sensitivity of different MU methods to the Avg.Gap with Retrain under varying amounts of forgotten data. \textbf{(b)} demonstrates the sensitivity of different MU methods to key hyperparameters in the forgetting mechanism with respect to Unlearning Ability (UA). \textbf{(c)} illustrates the sensitivity of different MU methods to key hyperparameters in the forgetting mechanism with respect to Generalization Ability (TA). \textbf{(d)} displays the unlearning curves for Retrain, GA, and CUFG.}\label{fig: Instability Analysis}
	\vspace{0.2cm}
\end{figure}
%------------------------------------------------------
%图a：展示了不同遗忘数据量对各MU方法在与Retrain的平均差距（Avg.Gap）上的敏感性。图b：展示了不同MU方法对遗忘机制中关键超参数在遗忘能力（UA）上的敏感性。图c：展示了不同MU方法对遗忘机制中关键超参数在泛化能力（TA）上的敏感性。图d：展示了Retrain、GA和CUFG在训练集上的Unlearning曲线。
\textbf{Instability Analysis of MU in Different standards.} Returning to the core Question (Q), %besides the effectiveness of MU methods, we focus on their impact on model stability within the unlearning mechanism. From the MU perspective, 
we aim for a robust approach that ensures model stability across all aspects after the forgetting task is completed, avoiding irreversible damage or increased uncertainty while maintaining efficient forgetting. Overall, the question can be concretized as the stability of MU methods and their impact on model stability, where model stability is closely tied to the stability of the forgetting mechanism. Only when the forgetting mechanism is stable and reliable will the impact on model stability be minimal.
% 回到本文的出发点：问题Q，除MU方法的有效性之外，MU方法在反学习过程中的稳定性是我们着重关注的事情。从MU的角度，我们不希望MU方法在高效遗忘的同时，对模型造成不可逆的伤害与不确定性增量。我们希望通过一种稳健的MU方法，在遗忘任务完成后，模型的各个方面仍维持基本稳定。总的来说，该问题可具象化的呈现为MU方法的稳定性以及在其影响下模型的稳定性。当然，模型的稳定性与遗忘机制本身的稳定性息息相关。若遗忘机制是稳定的、可信的，那么模型稳定性不会受到很大的冲击。具体地，我们通过分析MU面对不同遗忘数据量的性能、其遗忘机制在关键超参数影响下的敏感性，来检验MU方法是否是稳健的。其次，我们还检验了在所提出方法的反学习曲线，即反学习过程中在保留训练集数据上的准确率，用于检验反学习进程影响下模型的稳定性。
In this study, we assess the robustness of MU methods by analyzing their performance under varying amounts of forgotten data and sensitivity to key hyperparameters. We evaluate the unlearning curves of the proposed method, examining the impact of the unlearning process on model stability.

As shown in figure \ref{fig: Instability Analysis}(a), the performance gap (Avg.Gap) between CUFG and Retrain does not significantly widen as the forgotten data increases from 10\% to 50\%. Compared to other advanced baselines, CUFG consistently maintains the smallest gap. Figures \ref{fig: Instability Analysis}(b) and \ref{fig: Instability Analysis}(c) show the sensitivity of MU methods to key hyperparameters in forgetting and generalization performance. It is important to note that the hyperparameter ranges are kept within reasonable limits to prevent aggressive MU methods from causing model failures. From the figures, CUFG’s box plots are closest to Retrain in both metrics. Additionally, CUFG’s smaller box size compared to other baselines further confirms its stability, demonstrating superior robustness in both Unlearning ability (UA) and generalization ability (TA). Figure \ref{fig: Instability Analysis}(d) illustrates the model stability in CUFG’s unlearning process, along with comparisons to Retrain and GA’s unlearning curves. The enlarged subplot shows that CUFG’s curve stability closely matches Retrain, while GA’s aggressive approach significantly harms the model.
%如3(a)所示, 从10%到50%的遗忘数据增量并未显著性的大幅拉伸CUFG与Retrain之间的平均性能差距Avg.Gap。相较于其余先进的基线方法，CUFG的始终稳定地呈现最小差距。图3b和3c分别展示了MU方法对关键超参数在遗忘性能与泛化性能方面的敏感性。当然，这些关键参数的随即取值范围都被限制于合理范围内。否则，部分激进的MU方法容易造成模型崩溃的后果。首先，我们根据图中信息易知，CUFG的箱体位置在两个指标上均是最接近Retrain的标准的。此外，与其余基线方法在箱体大小方面的比较进一步验证了CUFG的稳定性。CUFG无论是遗忘性能UA还是泛化性能TA在敏感性分析中均优于现有基线方法。CUFG反学习进程中的模型稳定性由图4展示。图中同时还展示了Retrain和GA的反学习曲线。通过放大的子图我们可以发现，CUFG曲线的稳定性是显著逼近于Retrain的。而GA的激进处理显然对模型是有很大伤害的。
%\setlength{\intextsep}{2pt}
\begin{wrapfigure}{r}{0.5\textwidth}
	\setlength{\abovecaptionskip}{-0.5cm}
	\setlength{\belowcaptionskip}{0cm}
        \vspace{-0.3cm}
	\begin{center}
		\subfloat[Weight Similarity Heatmap]{			\includegraphics[width=0.245\textwidth, height=3.2cm]{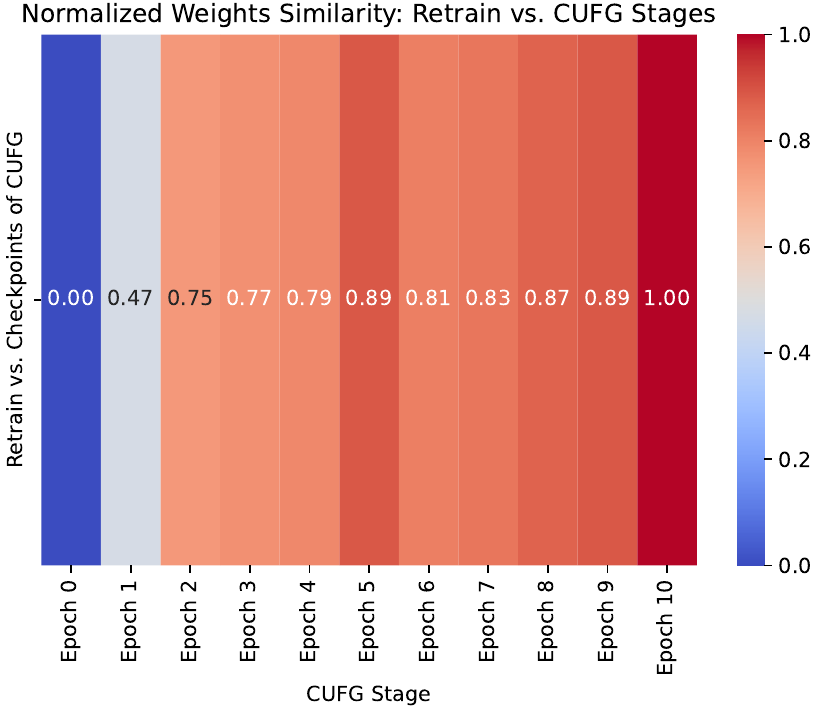}}
		\subfloat[Confidence Distribution]{
			\includegraphics[width=0.245\textwidth, height=3.2cm]{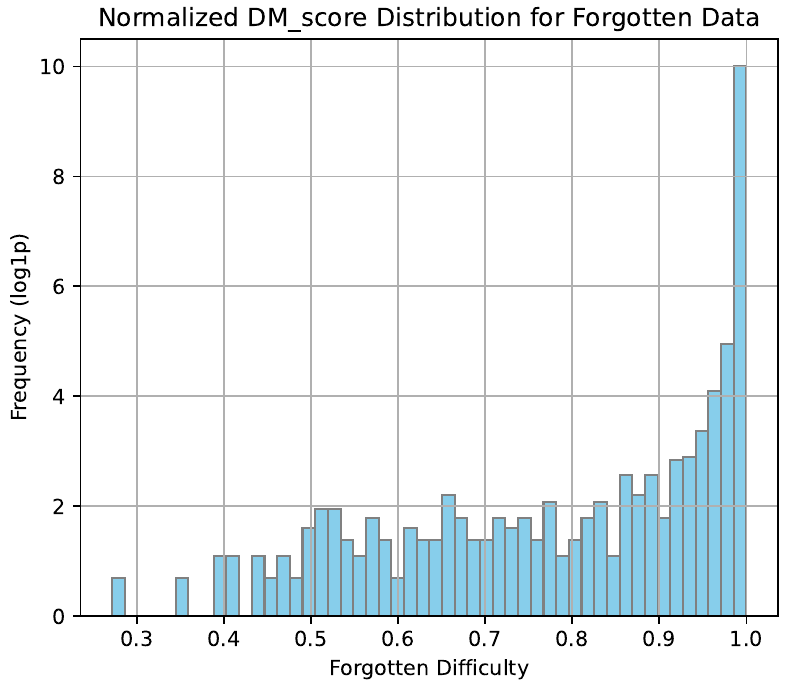}}
	\end{center}
        \vspace{0.5cm}
	\caption{Experimental validation results of CUFG motivation assumptions.}
        \label{fig:motivation}% {r} 表示右对齐，宽度40%文本宽度
\end{wrapfigure}
\textbf{Motivation Verification of CUFG.} The innovation of CUFG lies in its two key designs: the unlearning mechanism guided by the forgetting gradient and the curriculum unlearning strategy. These are grounded in two fundamental assumptions: (i) The gradient-corrector, guided by the forgetting gradient, helps the model progressively approach a local optimum of the MU optimization problem, as illustrated in Figure \ref{fig:1}(c); (ii) The trained model exhibits varying confidence levels towards forgotten data, leading to differences in forgetting difficulty. These assumptions are validated in Figure \ref{fig:motivation}. Figure \ref{fig:motivation}(a) shows a heatmap of the similarity between the model's weights during the unlearning process and the golden standard Retrain weights, indicating that our forgetting mechanism design is valid and aligned with our intentions. Figure \ref{fig:motivation}(b) displays the frequency distribution histogram of the model’s confidence in randomly forgotten data, showing that the forgetting difficulty varies across different data, further confirming the effectiveness of our curriculum unlearning strategy.
\begin{wraptable}{r}{0.6\textwidth}  % "r" 表示表格放在右侧，0.5 是表格宽度占页面宽度的比例
    \centering
    \caption{Comparison of MU efficiency under the forgetting scenario: random data forgetting (10\%) on ResNet-18}
    \vspace{0.2cm}
    \scalebox{0.66}{
    \begin{tabular}{@{}c|ccccccccc@{}}
\toprule [2pt]
Methods  & Retrain & FT   & GA   & IU   & BE   & BS   & L1-sparse & SalUn & CUFG \\ [2pt] \midrule
RTE(min) & 84.37   & 4.13 & 2.36 & 6.74 & 1.35 & 1.69 & 4.08      & 4.72  & 5.44 \\ [2pt] \bottomrule [2pt]
\end{tabular}}
\label{Efficiency}
\end{wraptable}
%CUFG的创新性集中于其受遗忘梯度指引的梯度修正遗忘机制与课程反学习数据遗忘策略两方面的设计。而它们是分别建立在两个基本假设上的。(i)受遗忘梯度指引的梯度策略能够指引模型如图1(c)所示的那样，逐步向MU优化问题的局部最优解逼近。(ii)已训练模型对于随机遗忘数据的置信度是不同的，因而对应遗忘难度亦有差异。这两个假设，我们在图4中进行了验证。图4(a)是模型在反学习过程中的权重与黄金标准Retrain权重的相似度热力图。其热力图趋势证明我们遗忘机制的设计是正确，并未违背初衷。图4(b)则是已训练模型对随机遗忘数据置信度的频率分布直方图。我们可以发现模型对不同数据的遗忘难度是不一致的。这证明我们对课程反学习遗忘策略的深入探索是有意义的。

\textbf{MU Efficiency.} As shown in Table \ref{Efficiency}, under the forgetting scenario of "Random data forgetting (10\%) on ResNet-18,"
we compare the MU efficiency of CUFG with other advanced baseline methods. The results show that CUFG is slightly less efficient than some baselines, mainly due to the time consumption of the forgetting gradient inference process. However, the efficiency gap is not significant, and CUFG still saves considerable computational resources compared to the golden standard Retrain. Given its performance advantages and contribution to model stability, the slight efficiency trade-off is acceptable.
%如表2所示，我们在遗忘场景：random data forgetting (10\%) on ResNet-18下，将CUFG与其余先进基线方法在MU效率方面进行了比较。实验数据表明，CUFG在MU效率方面略弱于部分基线方法。这是因为遗忘梯度推理的过程消耗了时间。令人欣慰的是，CUFG落后并不是很多，相较于黄金标准Retrain，还是可以节省很多的计算资源。我们认为以其性能优势与对模型稳定性方面的帮助，略微的效率牺牲是可以忽略的。

%----------------------------------------------
\begin{figure}[ht]
	\setlength{\abovecaptionskip}{-0.5cm}
	\setlength{\belowcaptionskip}{0cm}
	\begin{center}
		\subfloat[Retrain]{	 
           \includegraphics[width=0.16\textwidth, height=2.2cm]{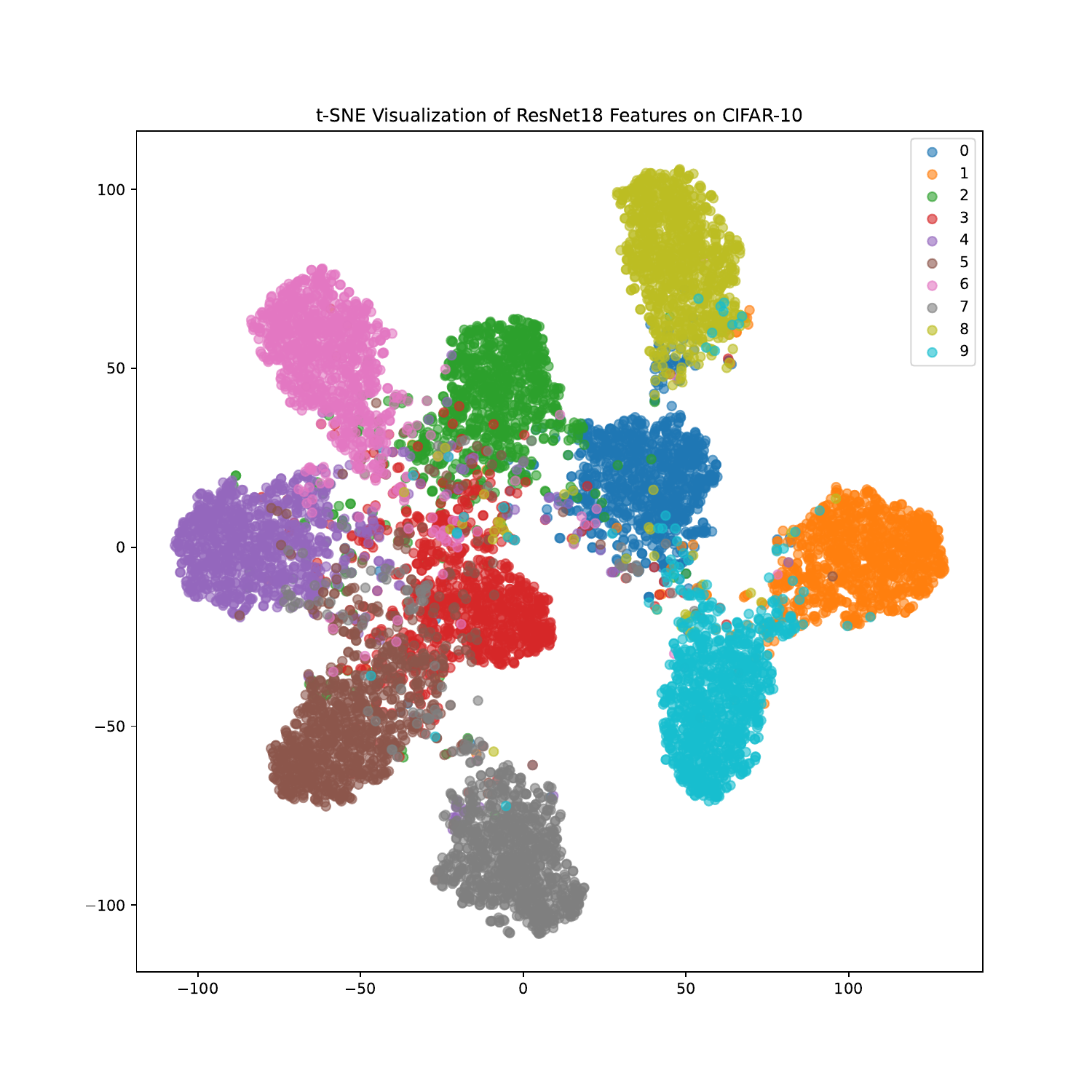}}
		\subfloat[FT]{
			\includegraphics[width=0.16\textwidth, height=2.2cm]{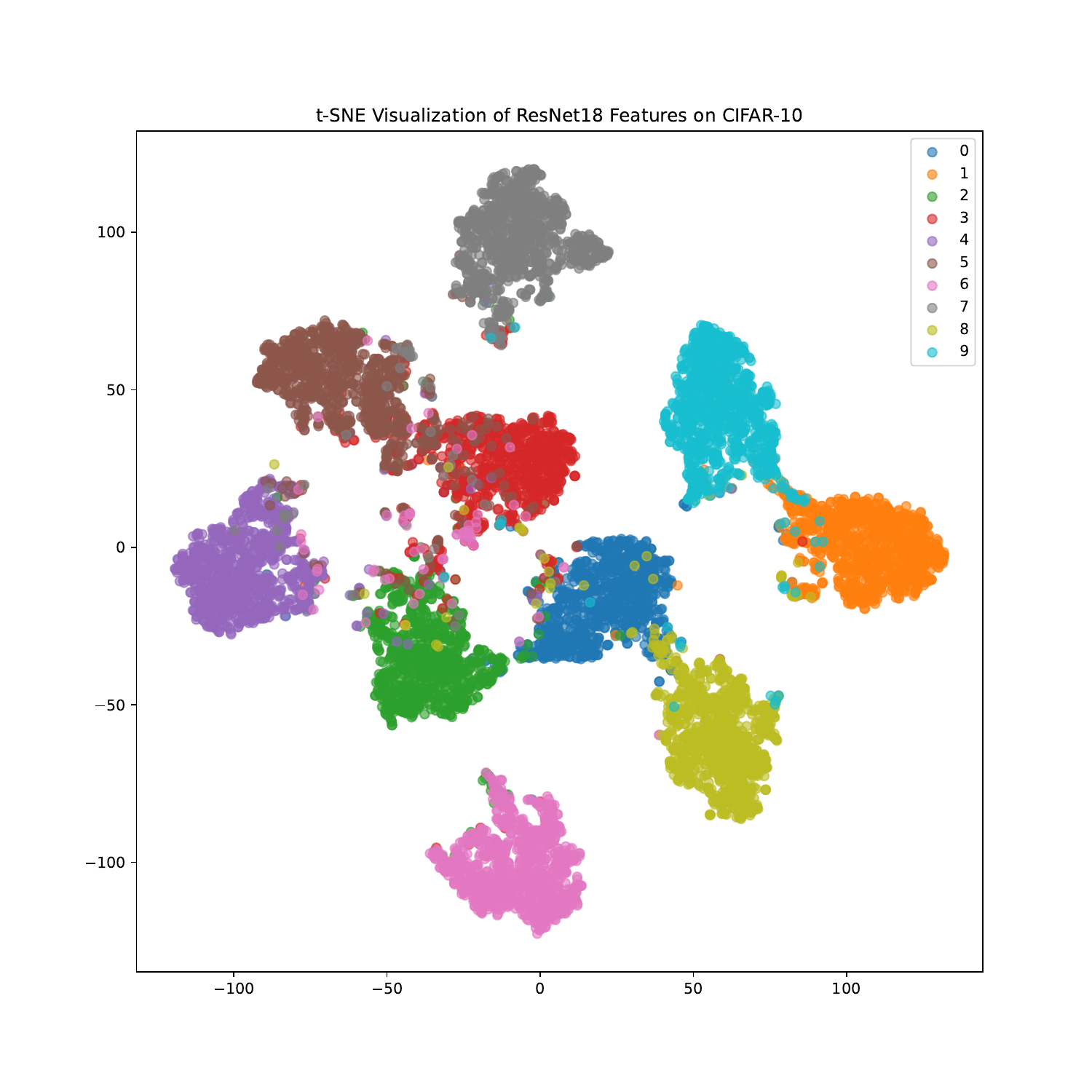}}
            \subfloat[GA]{
			\includegraphics[width=0.16\textwidth, height=2.2cm]{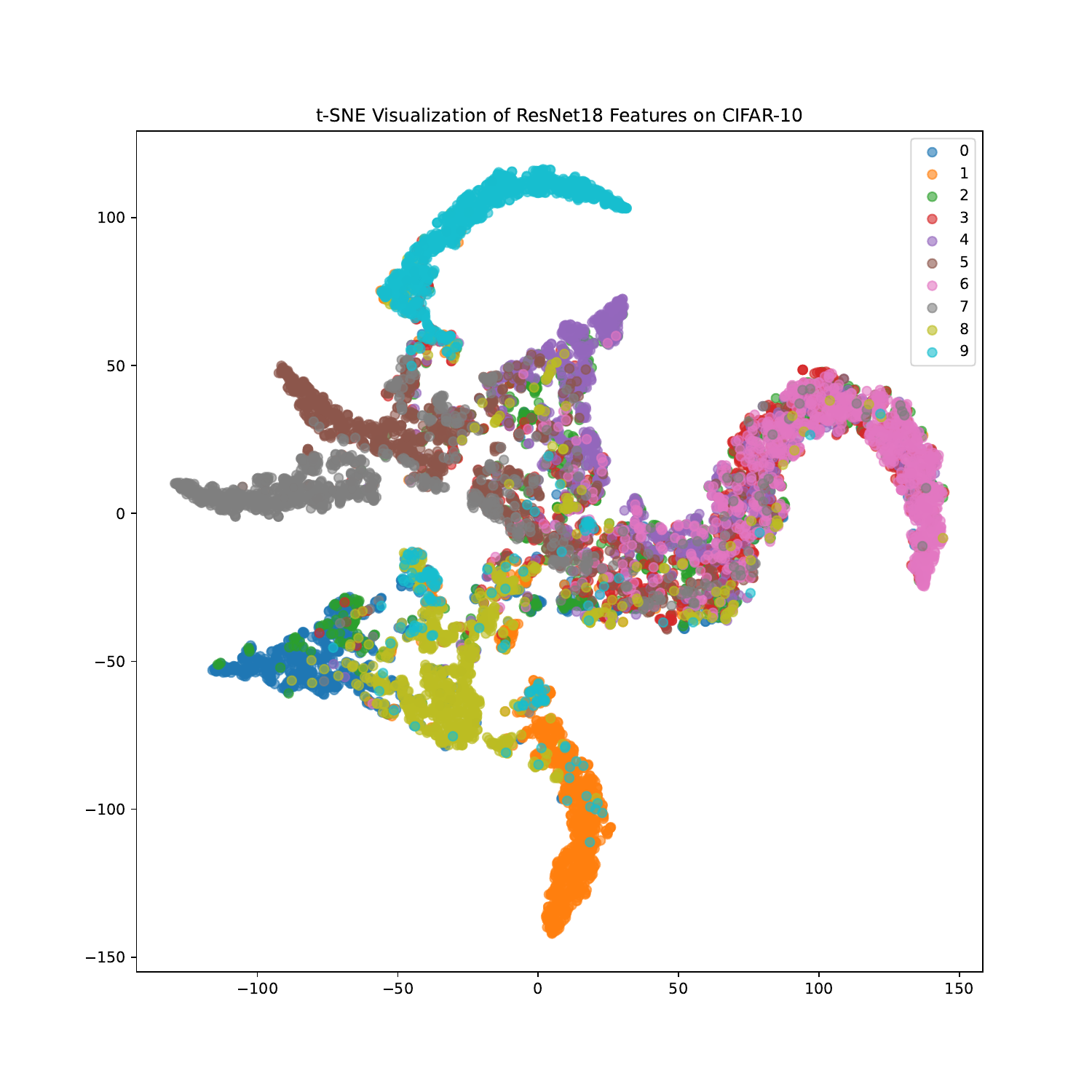}}
            \subfloat[L1-sparse]{
			\includegraphics[width=0.16\textwidth, height=2.2cm]{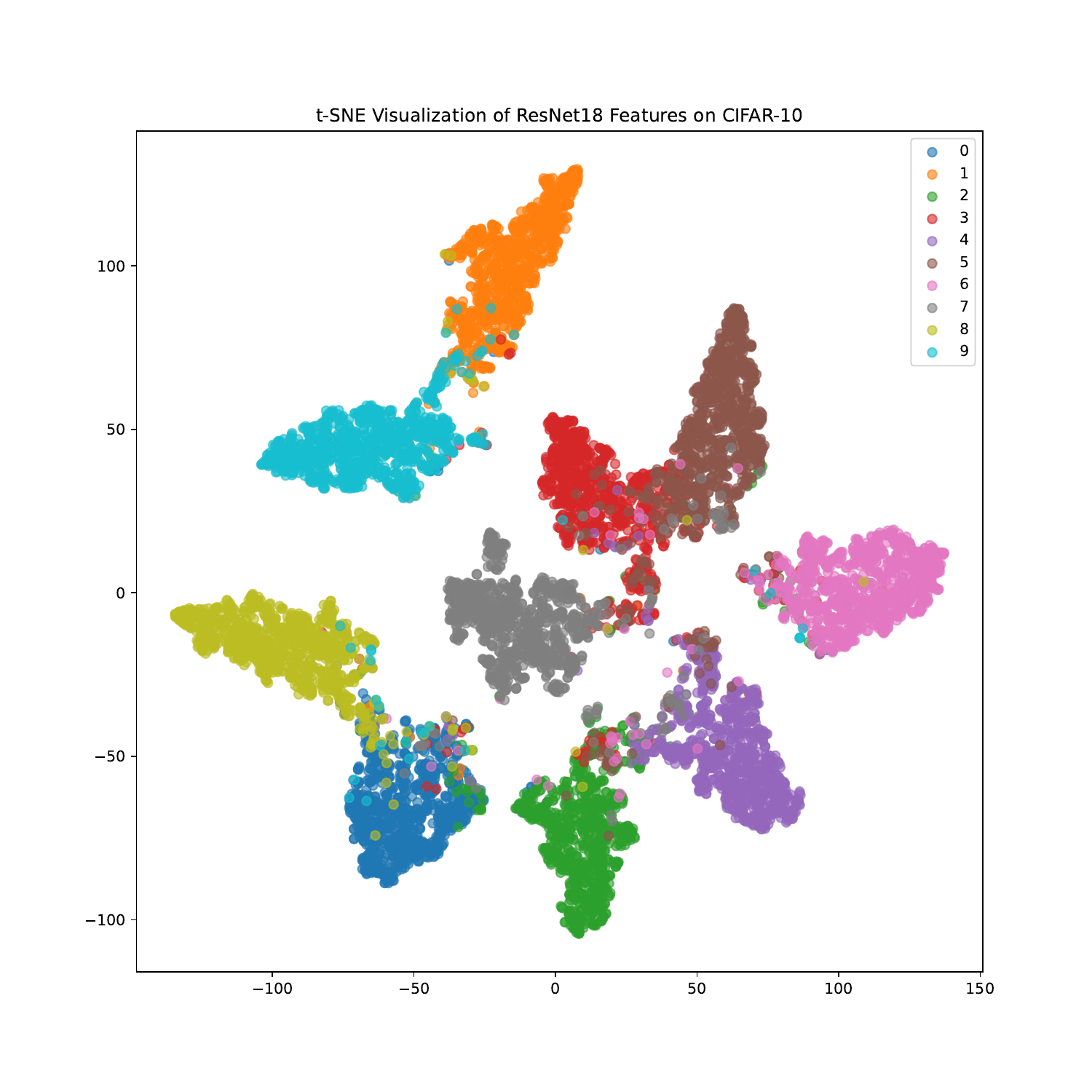}}
            \subfloat[SalUn]{
			\includegraphics[width=0.16\textwidth, height=2.2cm]{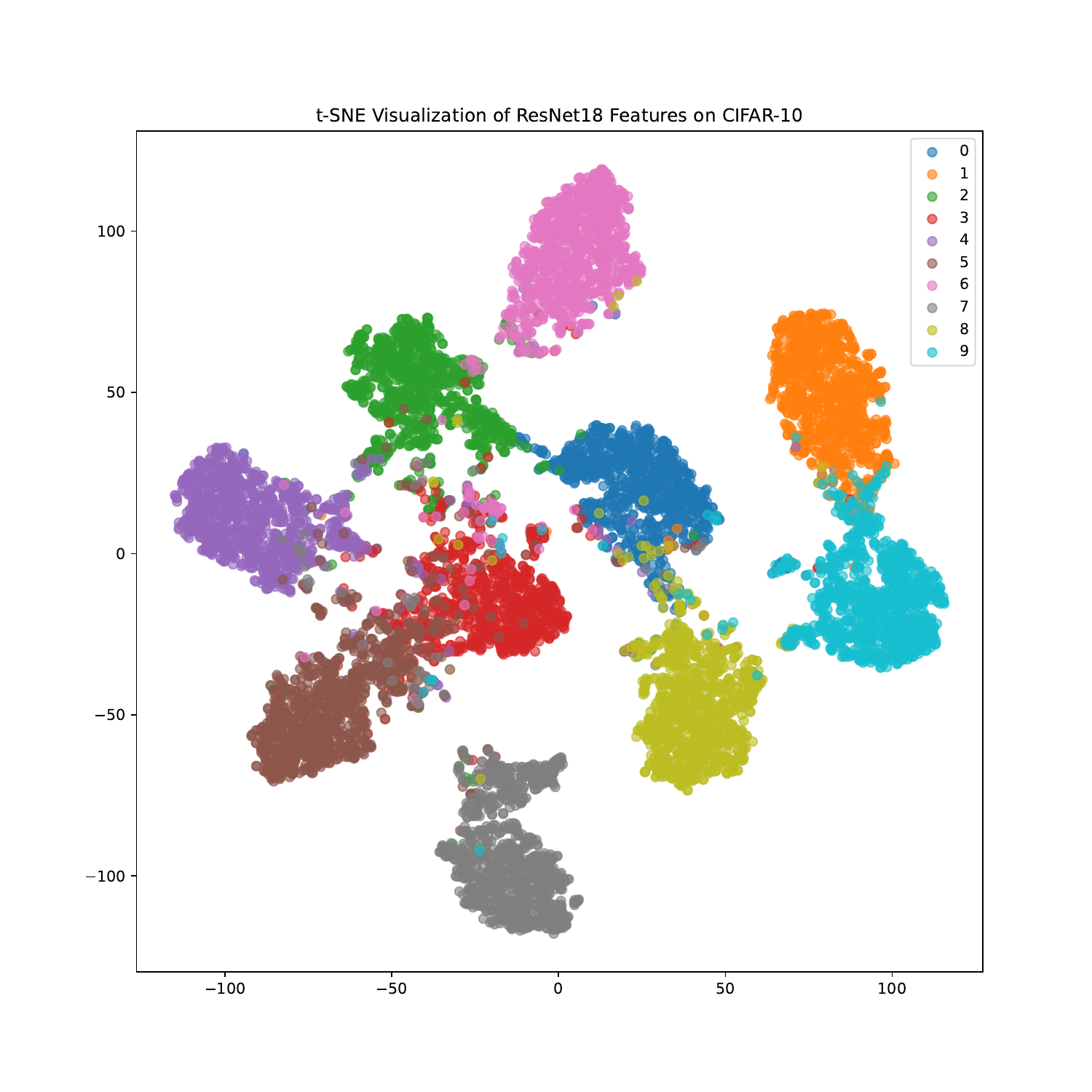}}
            \subfloat[CUFG]{
			\includegraphics[width=0.16\textwidth, height=2.2cm]{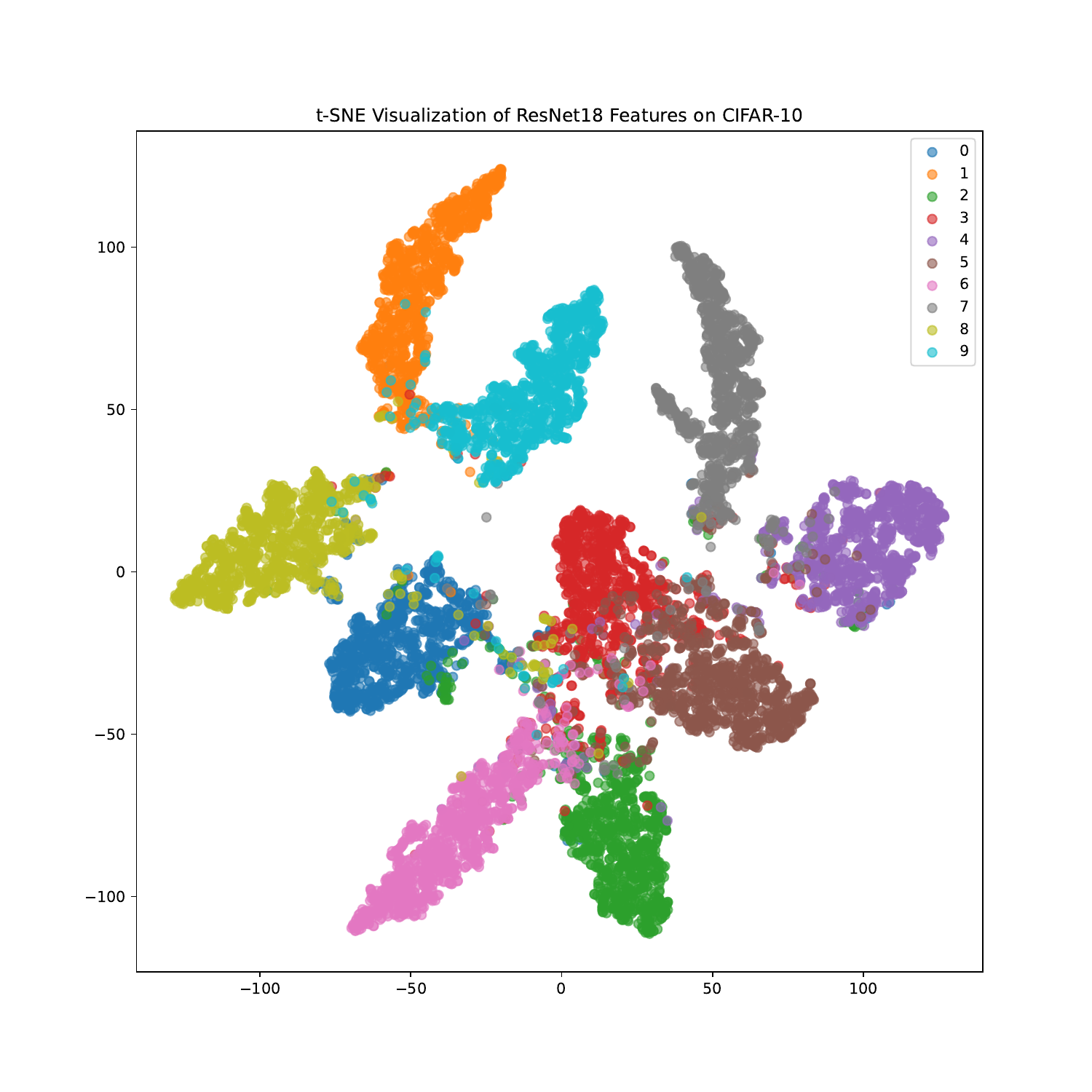}}
	\end{center}
        \vspace{0.5cm}
	\caption{Comparison of t-SNE visualization of model image classification after unlearning.}\label{fig: tsne}
	\vspace{0.2cm}
\end{figure}
%------------------------------------------------------
\textbf{Visualization Analysis.} Figure \ref{fig: tsne} shows the t-SNE visualization results after scrub forgetting data using various MU methods. First, CUFG exhibits better class separation and more compact intra-class clustering, indicating that our unlearning approach is robust and has minimal impact on model performance. Secondly, a detailed comparison of the zoomed-in regions with the most confusion shows that the misclassification distribution of the forgotten data is most similar to that of Retrain, validating the effectiveness of our strategy to progressively approach the local optimum.
%图5展示了CUFG与各基线方法Scrub遗忘数据后，在CIFAR-10数据集上的分类可视化结果。首先，CUFG呈现出更好的类分离效果和更紧凑的类内聚类，这说明我们的Unlearning方案是稳健的，并未对模型的性能产生较大的冲击。其次，放大图，仔细对比可以发现，我们遗忘的数据错误分类的混乱分布情况与Retrain的结果是最相近的。这说明我们逐步寻找局部最优解的策略是奏效的。

\section{Conclusion}
This paper presents two innovative approaches for non-aggressive, stable MU methods from the perspectives of forgetting mechanism design and data forgetting strategy. Inspired by the golden standard Retrain, the gradient correction method guided by the forgetting gradient steadily approaches the local optimum. Additionally, we propose the new concept of curriculum unlearning to address the inconsistency in data forgetting difficulty, enabling the model to forget from easy to hard. Extensive experiments validate the rationale and effectiveness of our approach. Notably, the curriculum unlearning paradigm is still in its early stages, with great research potential in subproblems like forgetting difficulty measurement and criteria generation, laying the foundation for more excellent MU methods in the future. 
%本文从遗忘机制设计与数据遗忘策略的全面角度出发，提出了对非激进式稳健MU方法的两项创新思考。受黄金标准Retrain启发，基于遗忘梯度的梯度修正方法逐步稳健地逼近局部最优解。同时，针对数据遗忘难度的不一致性，我们提出了课程反学习的创新理念，帮助模型实现由易到难的遗忘。大量实验验证了我们动机的合理性和创新的有效性。值得注意的是，课程反学习范式仍处于初步探索阶段，遗忘难度度量和课程划分等子问题有很大的研究潜力，为未来更稳健的MU方法奠定了基础。


%\bibliographystyle{IEEEtran}  % 或 abbrvnat, unsrtnat 都行
%\bibliography{IEEEabrv, Styles_2025/CUFG_NeurIPS25}     % 确保和你的 .bib 文件名一致
\bibliographystyle{plainnat}
\bibliography{main}

%%%%%%%%%%%%%%%%%%%%%%%%%%%%%%%%%%%%%%%%%%%%%%%%%%%%%%%%%%%%
%\iffalse
\newpage
\appendix

\section{ALGORITHM FRAMEWORKS OF UFG AND CUFG} \label{APP:A}
\scalebox{0.99}{
\begin{minipage}{\linewidth}
\begin{algorithm}[H]
\caption{The algorithm for UFG}\label{alg:alg1}
\KwIn{Trained model $h(\theta_D^*)$, retained dataset $D_r$, forget dataset $D_f$, Gradient correction threshold hyperparameter $\gamma$, learning rate $\eta$, and number of epochs $\mathbb{E}$. }
\KwOut{Model $h(\theta_U)$, in which the forgotten data has been scrubbed.}
Load trained model weights $\theta_U \longleftarrow \theta_D^*$ \;
\For{$\mathrm{epoch} \longleftarrow 0,1,\dots, \mathbb{E}-1$}{
    Reasoning about forgotten data: $\nabla \theta_{U_{(x_j,y_j)}} = \nabla_{\theta_{U}} \left( \mathcal{L}(h(x_j; \theta_U), y_j) \right)$\;
    Average Gradient: $- \nabla \theta_{U_{D_f}}^{m} = - \frac{1}{|{D_f}|} \sum_{(x_j, y_j) \in {D_f}} \nabla \theta_{U_{(x_j,y_j)}}$ \;
    Forgotten Gradient: $- \nabla \theta_{U_{D_f}}^{m}$ \;
    \For{$b \longleftarrow \mathrm{all \ batches \ of \ }  D_r$}{
        Fine-tuning gradient on $D_r$ of the current batch: $\nabla \theta_{U_{D_r}} = \nabla_{\theta_{U}} \left(\mathcal{L}(\theta; b) \mid _{\theta = \theta_U} \right)$ \;
        Calculate the gradient angle: $\angle \alpha = \mathrm{Angle}(\nabla \theta_{U_{D_r}}, \nabla \theta_{U_{D_f}}^{m})$\;
        \If{$\angle \alpha > \gamma$}{
            The unlearning gradient $\nabla \theta_{U}$ is $\nabla \theta_{U_{D_r}}$: $\nabla \theta_{U} \longleftarrow \nabla \theta_{U_{D_r}}$  \;
        }
        \Else{
            The unlearning gradient $\nabla \theta_{U}$ adjusted: $\nabla \theta_{U} \longleftarrow \frac{1}{2} (-\nabla \theta_{U_{D_f}}^{m} + \nabla \theta_{U_{D_r}})$ \;
        }
        One-step SGD updates model weights: $\theta_{U} \longleftarrow \theta_{U}-\nabla \theta_{U}$ \;
    }
}
\Return{\textnormal{The final weight after iterative optimization of the UFG algorithm} $\theta_{U}$.}
\end{algorithm}
\end{minipage}
}
%------------------------------------------------------------
\scalebox{0.99}{
\begin{minipage}{\linewidth}
\begin{algorithm}[H]
\caption{The algorithm for CUFG}\label{alg:alg2}
\KwIn{Trained model $h(\theta_D^*)$, retained dataset $D_r$, forget dataset $D_f$, Gradient correction threshold hyperparameter $\gamma$, learning rate $\eta$, and number of epochs $\mathbb{E}$. }
\KwOut{Model $h(\theta_U)$, in which the forgotten data has been scrubbed.}
Measure the difficulty of unlearning for each sample in $D_f$: $\mathrm{DM_{score}} = \mathcal{M}(h(\theta_D^*),D_f)$ \;
Divide $D_f$ according to $\mathrm{DM_{score}}$: $D_f^i =  \mathcal{I}(\left\{x_1, x_2, \dots, x_{\left | D_f \right |} \right\}, i), \quad i = 1, \dots, n$\;
Construct a series of MU criteria accordingly: $C_u = \langle T_1, \ldots, T_i, \ldots, T_n \rangle$ \;
Load trained model weights $\theta_U \longleftarrow \theta_D^*$ \;
\For{$T_i \longleftarrow T_1,T_2,\dots,T_n$}{
    \For{$\mathrm{epoch} \longleftarrow 0,1,\dots, \mathbb{E}/n-1$}{
        Reasoning about forgotten data of $D_f^i$: $\nabla \theta_{U_{(x_j,y_j)}} = \nabla_{\theta_{U}} \left( \mathcal{L}(h(x_j; \theta_U), y_j) \right)$\;
        Average Gradient: $- \nabla \theta_{U_{D_f^i}}^{m} = - \frac{1}{|{D_f^i}|} \sum_{(x_j, y_j) \in {D_f^i}} \nabla \theta_{U_{(x_j,y_j)}}$ \;
        Forgotten Gradient: $- \nabla \theta_{U_{D_f^i}}^{m}$ \;
        \For{$b \longleftarrow \mathrm{all \ batches \ of \ }  D_r$}{
            Fine-tuning gradient on $D_r$ of the current batch: $\nabla \theta_{U_{D_r}} = \nabla_{\theta_{U}} \left(\mathcal{L}(\theta; b) \mid _{\theta = \theta_U} \right)$ \;
            Calculate the gradient angle: $\angle \alpha = \mathrm{Angle}(\nabla \theta_{U_{D_r}}, \nabla \theta_{U_{D_f^i}}^{m})$\;
            \If{$\angle \alpha > \gamma$}{
                The unlearning gradient $\nabla \theta_{U}$ is $\nabla \theta_{U_{D_r}}$: $\nabla \theta_{U} \longleftarrow \nabla \theta_{U_{D_r}}$  \;
            }
            \Else{
                The unlearning gradient $\nabla \theta_{U}$ adjusted: $\nabla \theta_{U} \longleftarrow \frac{1}{2} (-\nabla \theta_{U_{D_f^i}}^{m} + \nabla \theta_{U_{D_r}})$ \;
            }
            One-step SGD updates model weights: $\theta_{U} \longleftarrow \theta_{U}-\nabla \theta_{U}$ \;
        }
    }
}
\Return{\textnormal{The final weight after iterative optimization of the CUFG algorithm} $\theta_{U}$.}
\end{algorithm}
\end{minipage}
}

\section{ADDITIONAL IMPLEMENTATION DETAILS} \label{APP:B}
\subsection{DATASETS AND MODELS} \label{APP:B1}
%我们在实验中对于数据集以及 backbone 模型的实验设置均遵从与 L1-Sparse 和 SALUN 方法相同的设定，以确保公平对比和可复现性。具体而言，我们在所有实验中统一使用了CIFAR-10、CIFAR-100，以及SVHN数据集，并采用了与原方法一致的划分方式与预处理流程。此外，在模型架构方面，我们选用了与原方法中一致的主干网络结构，ResNet-18和VGG-16，并保持其初始化方式、训练轮数、优化器参数（如学习率、权重衰减）等关键超参数一致。这样的设置确保了我们方法的性能提升能够归因于算法本身的改进，而非由于不同的实验条件所带来的偏差。
In our experiments, we followed the same settings for datasets and backbone models as those used in the L1-sparse \cite{jia2023model} and SalUn \cite{fan2024salun} methods to ensure fair comparison and reproducibility. Specifically, we consistently used the CIFAR-10, CIFAR-100, and SVHN datasets across all experiments, adopting the same data splits and preprocessing procedures as in the original methods. Furthermore, regarding model architecture, we employed the same backbone networks—ResNet-18 and VGG-16—as used in the original works, maintaining identical initialization schemes, number of training epochs, and optimizer configurations (e.g., learning rate, weight decay). These consistent settings ensure that any performance improvement of our method can be attributed to algorithmic advances rather than differences in experimental conditions.

\subsection{ADDITIONAl TRAINING AND UNLEARNING SETTINGS} \label{APP:B2}
%对于除GA之外的所有方法，我们均采用10个epochs进行unlearning。所有的baselines以及我们所提出的UFG与CUFG均可以分为微调式，非微调式的方法。微调式的unlearning学习率均设置为0.01，而非微调式则根据原文所给出的参数区间进行选取。本文默认SGD的优化器选择。为实现公平的对比，随机数据遗忘以及随机类遗忘两遗忘场景的设置时，随机种子选取是一致的。在稳定性实验部分，我们敏感性分析实验所选取的超参数均为所有遗忘方法在遗忘机制设计上最相关的超参数，比如GA的学习率，L1-Sparse的权重稀疏度，以及SalUn制造权重mask时所选取的阈值。对于我们的方法来说，我们选择的时角度阈值。遗忘效率方面，各方法运行时，计算机设备处于公平环境。
Except for the GA method, all unlearning methods were conducted using 10 epochs during the unlearning process. All baseline methods, including our proposed UFG and CUFG, can be categorized into two types: fine-tuning-based and non-fine-tuning-based approaches. For fine-tuning-based methods, the learning rate during unlearning was uniformly set to 0.01, whereas for non-fine-tuning-based methods, the hyperparameters were selected according to the ranges suggested in their original papers. The SGD optimizer was used by default across all experiments.

To ensure fair comparison, we fixed the random seed across all experiments in both the random sample forgetting and random class forgetting scenarios, so that each method was evaluated under the same initialization conditions. For the model stability evaluation, we conducted a sensitivity analysis to investigate the robustness of each method to variations in its key hyperparameters. The selected hyperparameters were chosen based on their central role in the design of each unlearning mechanism—for example, the learning rate in GA \cite{graves2021amnesiac, thudi2022unrolling}, the sparsity coefficient in L1-sparse \cite{jia2023model}, and the threshold used to generate the weight mask in SalUn \cite{fan2024salun}. For our proposed methods, UFG and CUFG, we focused on analyzing the angular threshold parameter $\gamma$. Regarding unlearning efficiency (i.e., RTE), all methods were executed under the same computational environment to eliminate differences due to hardware performance. This includes using the same hardware platform, identical GPU configuration, and consistent system load to ensure fair and comparable results.

\subsection{IMPLEMENTATION DETAILS OF THE MIA METRIC} \label{APP:B3}
\textbf{Membership Inference Attack (MIA)} aims to determine whether a specific data point was part of a model’s training set by analyzing its output predictions and confidence scores. The key assumption is that models behave differently on training versus unseen data, which can be exploited for inference.

MIA involves two stages: training and testing. In the training stage, a balanced dataset is created from the remaining data ($D_r$) and an unrelated test set (different from the forgetting dataset $D_f$), and is used to train an MIA classifier to distinguish members from non-members. This balance is essential to avoid bias. In the testing stage, the trained MIA classifier is applied to the unlearned model ($h(\theta_U)$) to evaluate whether it correctly identifies $D_f$ samples as non-members. In other words, we use privacy attacks to help us judge the quality of the forgetting performance of the MU methods.

\subsection{EXPERIMENTAL CONFIGURATION AND COMPUTING RESOURCES} \label{APP:B4}
The experiments involving UFG, CUFG, and their comparisons were conducted in a virtual environment using Python 3.9.18 and the deep learning framework PyTorch 2.0.1. To ensure efficiency and stability, all computational tasks were executed on a server equipped with an NVIDIA GeForce RTX 4090 GPU. This server is configured with 24GB of memory and high processing power, capable of handling large-scale datasets and complex model training. The operating system running on the server is Ubuntu 22.04.2 LTS, ensuring compatibility and long-term support for the development environment. Additionally, to further improve the reproducibility and stability of the experiments, we installed all necessary dependencies in the virtual environment and managed them with version control, ensuring consistent experimental conditions across trials.

\section{ADDITIONAL EXPERIMENT RESULTS} \label{APP:C}
%本小节我们补充了正文中放不下的实验。具体地，除正文中table1中展示的4种遗忘场景外，在CIFAR10、CIFAR100、SVHN数据集上，ResNet-18、VGG-16的backbone上还设定了其余的8种遗忘场景，如tableA1-A8所示。
In this subsection, we present additional experimental results that could not be included in the main text due to space limitations. Specifically, %beyond the four primary unlearning scenarios shown in Table \ref{performance} of the main paper, 
we further constructed and evaluated eight additional unlearning scenarios on the CIFAR-10, CIFAR-100, and SVHN datasets using ResNet-18 and VGG-16 as backbone architectures. The detailed configurations and results are provided in Tables \ref{A.1} to \ref{A.8}. These results further demonstrate that our UFG and CUFG methods consistently exhibit strong generalization ability and stable unlearning performance, even under variations in data distribution, class granularity, and model architecture.

Based on the information provided in Tables \ref{A.1} to \ref{A.8}, we present the following additional experimental analysis. First, the findings and advantages of the proposed methods discussed in the main text are not results obtained by chance. All experiments presented in this paper are verifiable. Second, using the controlled variables, we find that changes in data distribution do not affect the advantages of UFG and CUFG. Specifically, CUFG consistently achieves the smallest average gap across all metrics when compared to the gold standard, Retrain, on all datasets. UFG also maintains competitive performance throughout. Finally, a qualitative analysis of all results across different scenarios reveals that our methods consistently perform well in UA/MIA, while also showing strong results in RA/TA. This indicates that the proposed general methods do not cause undue harm to the model during the unlearning process, confirming their significant advantage in terms of stability.

%此外，我们根据TableA1-A8提供的信息，补充如下实验分析：首先，正文中所阐述的实验分析与所提方法的性能优势均非偶然情况下的实验结果。本文的全部实验均可对应验证。其次，根据控制变量法，控制其他变量的情况下，数据分布的变化并未影响到UFG与CUFG的性能优势。CUFG在展示的全部数据集上实现了与黄金标准Retrain在所有指标上最小平均差距。UFG亦始终保持优秀的竞争力。最后，从不同遗忘场景的所有实验结果定性分析，UFG与CUFG在保证遗忘性能（UA\MIA）的前提下，在RA和TA的表现上始终表现良好。这表明本文提出的通用方法在遗忘进程中并未过度损伤模型本身，说明在稳定性方面确实具备显著优势。

\vspace{0.22cm}

\renewcommand{\thetable}{A.1} 
\begin{table}[htbp]
\centering
\caption{Comparison of MU performance among various methods on the CIFAR-100 dataset using ResNet-18 as the backbone network, under the forgetting scenario: Random Data Forgetting 10\%.}
\label{A.1}
\begin{tabular}{@{}p{2cm}<{\centering}|p{2cm}<{\centering}p{2cm}<{\centering}p{2cm}<{\centering}p{2cm}<{\centering}p{2cm}<{\centering}@{}}
\toprule [2pt]
\multirow{2}{*}{\textbf{Methods}} & \multicolumn{5}{c}{\textbf{Random Data Forgetting (10\%) on CIFAR-100 with ResNet-18}}                                                                                                \\ [2pt]
                         & \multicolumn{1}{c|}{\textbf{UA}} & \multicolumn{1}{c|}{\textbf{RA}} & \multicolumn{1}{c|}{\textbf{TA}} & \multicolumn{1}{c|}{\textbf{MIA}}   & \textbf{Avg.Gap} \\ [2pt] \midrule [1pt]
Retrain                  & 22.73 (\textcolor{blue}{0.00})                           & 99.97 (\textcolor{blue}{0.00})                           & 74.10 (\textcolor{blue}{0.00})                           & \multicolumn{1}{p{2cm}<{\centering}|}{48.87 (\textcolor{blue}{0.00})}          & \textcolor{blue}{0.00}            \\ [2pt]
FT                 & 2.47 (\textcolor{blue}{20.26})                           & 99.95 (\textcolor{blue}{0.02})                           & 75.46 (\textcolor{blue}{1.36})                           & \multicolumn{1}{c|}{11.40 (\textcolor{blue}{37.47})}           & \textcolor{blue}{14.78}            \\ [2pt]
GA               & 2.58 (\textcolor{blue}{20.15})                           & 97.52 (\textcolor{blue}{2.45})                           & 75.41 (\textcolor{blue}{1.31})                           & \multicolumn{1}{c|}{6.02 (\textcolor{blue}{42.85})}           & \textcolor{blue}{16.69}             \\ [2pt]
IU              & 3.02 (\textcolor{blue}{19.71})                           & 97.22 (\textcolor{blue}{2.75})                           & 74.05 (\textcolor{blue}{0.05})                           & \multicolumn{1}{c|}{9.80 (\textcolor{blue}{39.07})}           & \textcolor{blue}{15.40}             \\ [2pt]
BE               & 2.41 (\textcolor{blue}{20.32})                           & 97.14 (\textcolor{blue}{2.83})                           & 73.95 (\textcolor{blue}{0.15})                           & \multicolumn{1}{c|}{9.75 (\textcolor{blue}{39.12})}           & \textcolor{blue}{15.61}             \\ [2pt]
BS               & 2.12 (\textcolor{blue}{20.61})                           & 97.32 (\textcolor{blue}{2.65})                           & 75.64 (\textcolor{blue}{1.54})                           & \multicolumn{1}{c|}{5.78 (\textcolor{blue}{43.09})}           & \textcolor{blue}{16.97}             \\ [2pt]
L1-sparse                & 11.73 (\textcolor{blue}{11.00})                           & 96.12 (\textcolor{blue}{3.85})                           & 69.69 (\textcolor{blue}{4.41})                           & \multicolumn{1}{c|}{21.51 (\textcolor{blue}{27.36})}          & \textcolor{blue}{11.66}             \\ [2pt]
SalUn                    & 24.40 (\textcolor{blue}{1.67})                           & 98.74 (\textcolor{blue}{1.23})                           & 69.13 (\textcolor{blue}{4.97})                           & \multicolumn{1}{c|}{75.09 (\textcolor{blue}{26.22})}          & \textcolor{blue}{8.52}             \\ [2pt] \midrule [1pt]
\textbf{UFG}         & \textbf{20.69 (\textcolor{blue}{2.04})}                   & \textbf{96.76 (\textcolor{blue}{3.21})}                   & \textbf{68.66 (\textcolor{blue}{5.44})}                   & \multicolumn{1}{c|}{\textbf{30.22 (\textcolor{blue}{18.65})}} & \textbf{\textcolor{blue}{7.34}}    \\ [2pt]
\textbf{CUFG}     & \textbf{20.93 (\textcolor{blue}{1.80})}                   & \textbf{98.39 (\textcolor{blue}{1.58})}                   & \textbf{69.41 (\textcolor{blue}{4.69})}                   & \multicolumn{1}{c|}{\textbf{29.04 (\textcolor{blue}{19.83})}} & \textbf{\textcolor{blue}{6.98}}    \\ [2pt] \bottomrule [2pt]
\end{tabular}
\end{table}

\vspace{0.22cm}

\renewcommand{\thetable}{A.2}
\begin{table}[htbp]
\centering
\caption{Comparison of MU performance among various methods on the CIFAR-100 dataset using ResNet-18 as the backbone network, under the forgetting scenario: Random Data Forgetting 50\%.}
\begin{tabular}{@{}p{2cm}<{\centering}|p{2cm}<{\centering}p{2cm}<{\centering}p{2cm}<{\centering}p{2cm}<{\centering}p{2cm}<{\centering}@{}}
\toprule [2pt]
\multirow{2}{*}{\textbf{Methods}} & \multicolumn{5}{c}{\textbf{Random Data Forgetting (50\%) on CIFAR-100 with ResNet-18}}                                                                                                \\ [2pt]
                         & \multicolumn{1}{c|}{\textbf{UA}} & \multicolumn{1}{c|}{\textbf{RA}} & \multicolumn{1}{c|}{\textbf{TA}} & \multicolumn{1}{c|}{\textbf{MIA}}   & \textbf{Avg.Gap} \\ [2pt] \midrule [1pt]
Retrain                  & 33.28 (\textcolor{blue}{0.00})                           & 99.98 (\textcolor{blue}{0.00})                           & 67.15 (\textcolor{blue}{0.00})                           & \multicolumn{1}{p{2cm}<{\centering}|}{60.96 (\textcolor{blue}{0.00})}          & \textcolor{blue}{0.00}            \\ [2pt]
FT                 & 2.68 (\textcolor{blue}{30.60})                           & 99.98 (\textcolor{blue}{0.00})                           & 75.67 (\textcolor{blue}{8.52})                           & \multicolumn{1}{c|}{10.76 (\textcolor{blue}{50.20})}           & \textcolor{blue}{22.33}            \\ [2pt]
GA               & 2.78 (\textcolor{blue}{30.50})                           & 97.48 (\textcolor{blue}{2.50})                           & 75.36 (\textcolor{blue}{8.21})                           & \multicolumn{1}{c|}{6.14 (\textcolor{blue}{54.82})}           & \textcolor{blue}{24.01}             \\ [2pt]
IU              & 13.36 (\textcolor{blue}{19.92})                           & 86.73 (\textcolor{blue}{13.25})                           & 62.49 (\textcolor{blue}{4.66})                           & \multicolumn{1}{c|}{17.64 (\textcolor{blue}{43.32})}           & \textcolor{blue}{20.29}             \\ [2pt]
BE               & 2.54 (\textcolor{blue}{30.74})                           & 97.50 (\textcolor{blue}{2.48})                           & 74.13 (\textcolor{blue}{6.98})                           & \multicolumn{1}{c|}{9.03 (\textcolor{blue}{51.93})}           & \textcolor{blue}{23.03}             \\ [2pt]
BS               & 3.11 (\textcolor{blue}{30.17})                           & 97.07 (\textcolor{blue}{2.91})                           & 73.34 (\textcolor{blue}{6.19})                           & \multicolumn{1}{c|}{8.83 (\textcolor{blue}{52.13})}           & \textcolor{blue}{22.85}             \\ [2pt]
L1-sparse                & 21.86 (\textcolor{blue}{11.42})                           & 91.94 (\textcolor{blue}{8.04})                           & 65.68 (\textcolor{blue}{1.47})                           & \multicolumn{1}{c|}{31.55 (\textcolor{blue}{29.41})}          & \textcolor{blue}{12.59}             \\ [2pt]
SalUn                    & 30.54 (\textcolor{blue}{2.74})                           & 91.36 (\textcolor{blue}{8.62})                           & 52.68 (\textcolor{blue}{14.47})                           & \multicolumn{1}{c|}{47.80 (\textcolor{blue}{13.16})}          & \textcolor{blue}{9.75}             \\ [2pt] \midrule [1pt]
\textbf{UFG}         & \textbf{33.69 (\textcolor{blue}{0.41})}                   & \textbf{93.77 (\textcolor{blue}{6.21})}                   & \textbf{61.27 (\textcolor{blue}{5.88})}                   & \multicolumn{1}{c|}{\textbf{38.70 (\textcolor{blue}{22.26})}} & \textbf{\textcolor{blue}{8.69}}    \\ [2pt]
\textbf{CUFG}     & \textbf{30.64 (\textcolor{blue}{2.64})}                   & \textbf{96.62 (\textcolor{blue}{3.36})}                   & \textbf{63.56 (\textcolor{blue}{3.59})}                   & \multicolumn{1}{c|}{\textbf{37.78 (\textcolor{blue}{23.18})}} & \textbf{\textcolor{blue}{8.19}}    \\ [2pt] \bottomrule [2pt]
\end{tabular}
\end{table}

\vspace{0.01cm}

\renewcommand{\thetable}{A.3}
\begin{table}[htbp]
\centering
\caption{Comparison of MU performance among various methods on the SVHN dataset using ResNet-18 as the backbone network, under the forgetting scenario: Random Data Forgetting 10\%.}
\begin{tabular}{@{}p{2cm}<{\centering}|p{2cm}<{\centering}p{2cm}<{\centering}p{2cm}<{\centering}p{2cm}<{\centering}p{2cm}<{\centering}@{}}
\toprule [2pt]
\multirow{2}{*}{\textbf{Methods}} & \multicolumn{5}{c}{\textbf{Random Data Forgetting (10\%) on SVHN with ResNet-18}}                                                                                                \\ [2pt]
                         & \multicolumn{1}{c|}{\textbf{UA}} & \multicolumn{1}{c|}{\textbf{RA}} & \multicolumn{1}{c|}{\textbf{TA}} & \multicolumn{1}{c|}{\textbf{MIA}}   & \textbf{Avg.Gap} \\ [2pt] \midrule [1pt]
Retrain                  & 4.63 (\textcolor{blue}{0.00})                           & 100.00 (\textcolor{blue}{0.00})                           & 95.72 (\textcolor{blue}{0.00})                           & \multicolumn{1}{p{2cm}<{\centering}|}{14.36 (\textcolor{blue}{0.00})}          & \textcolor{blue}{0.00}            \\ [2pt]
FT                 & 0.53 (\textcolor{blue}{4.10})                           & 100.00 (\textcolor{blue}{0.00})                           & 95.81 (\textcolor{blue}{0.09})                           & \multicolumn{1}{c|}{2.31 (\textcolor{blue}{12.05})}           & \textcolor{blue}{4.06}            \\ [2pt]
GA               & 0.64 (\textcolor{blue}{3.99})                           & 99.56 (\textcolor{blue}{0.44})                           & 95.67 (\textcolor{blue}{0.05})                           & \multicolumn{1}{c|}{1.27 (\textcolor{blue}{13.09})}           & \textcolor{blue}{4.39}             \\ [2pt]
IU              & 1.76 (\textcolor{blue}{2.87})                           & 98.72 (\textcolor{blue}{1.28})                           & 92.79 (\textcolor{blue}{2.93})                           & \multicolumn{1}{c|}{6.01 (\textcolor{blue}{8.35})}           & \textcolor{blue}{3.86}             \\ [2pt]
BE               & 0.42 (\textcolor{blue}{4.21})                           & 99.46 (\textcolor{blue}{0.54})                           & 95.23 (\textcolor{blue}{0.49})                           & \multicolumn{1}{c|}{1.16 (\textcolor{blue}{13.20})}           & \textcolor{blue}{4.61}             \\ [2pt]
BS               & 0.42 (\textcolor{blue}{4.21})                           &  99.45 (\textcolor{blue}{0.55})                           & 95.37 (\textcolor{blue}{0.36})                           & \multicolumn{1}{c|}{1.13 (\textcolor{blue}{13.23})}           & \textcolor{blue}{4.59}             \\ [2pt]
L1-sparse                & 4.19 (\textcolor{blue}{0.44})                           & 99.72 (\textcolor{blue}{0.28})                           & 94.02 (\textcolor{blue}{1.70})                           & \multicolumn{1}{c|}{9.61 (\textcolor{blue}{4.75})}          & \textcolor{blue}{1.79}             \\ [2pt]
SalUn                    & 6.55 (\textcolor{blue}{1.92})                           & 97.53 (\textcolor{blue}{2.47})                           & 94.24 (\textcolor{blue}{1.48})                           & \multicolumn{1}{c|}{15.62 (\textcolor{blue}{1.26})}          & \textcolor{blue}{1.78}             \\ [2pt] \midrule [1pt]
\textbf{UFG}         & \textbf{4.64 (\textcolor{blue}{0.01})}                   & \textbf{98.07 (\textcolor{blue}{1.93})}                   & \textbf{92.04 (\textcolor{blue}{3.68})}                   & \multicolumn{1}{c|}{\textbf{11.01 (\textcolor{blue}{3.35})}} & \textbf{\textcolor{blue}{2.24}}    \\ [2pt]
\textbf{CUFG}     & \textbf{4.64 (\textcolor{blue}{0.01})}                   & \textbf{99.34 (\textcolor{blue}{0.66})}                   & \textbf{93.42 (\textcolor{blue}{2.30})}                   & \multicolumn{1}{c|}{\textbf{10.52 (\textcolor{blue}{3.84})}} & \textbf{\textcolor{blue}{1.70}}    \\ [2pt] \bottomrule [2pt]
\end{tabular}
\end{table}

\vspace{0.01cm}

\renewcommand{\thetable}{A.4}
\begin{table}[htbp]
\centering
\caption{Comparison of MU performance among various methods on the SVHN dataset using ResNet-18 as the backbone network, under the forgetting scenario: Random Data Forgetting 50\%.}
\begin{tabular}{@{}p{2cm}<{\centering}|p{2cm}<{\centering}p{2cm}<{\centering}p{2cm}<{\centering}p{2cm}<{\centering}p{2cm}<{\centering}@{}}
\toprule [2pt]
\multirow{2}{*}{\textbf{Methods}} & \multicolumn{5}{c}{\textbf{Random Data Forgetting (50\%) on SVHN with ResNet-18}}                                                                                                \\ [2pt]
                         & \multicolumn{1}{c|}{\textbf{UA}} & \multicolumn{1}{c|}{\textbf{RA}} & \multicolumn{1}{c|}{\textbf{TA}} & \multicolumn{1}{c|}{\textbf{MIA}}   & \textbf{Avg.Gap} \\ [2pt] \midrule [1pt]
Retrain                  & 5.90 (\textcolor{blue}{0.00})                           & 100.00 (\textcolor{blue}{0.00})                           & 94.58 (\textcolor{blue}{0.00})                           & \multicolumn{1}{p{2cm}<{\centering}|}{19.30 (\textcolor{blue}{0.00})}          & \textcolor{blue}{0.00}            \\ [2pt]
FT                 & 0.45 (\textcolor{blue}{5.45})                           & 100.00 (\textcolor{blue}{0.00})                           & 95.75 (\textcolor{blue}{1.17})                           & \multicolumn{1}{c|}{2.26 (\textcolor{blue}{17.04})}           & \textcolor{blue}{5.92}            \\ [2pt]
GA               & 0.63 (\textcolor{blue}{5.27})                           & 99.54 (\textcolor{blue}{0.46})                           & 95.58 (\textcolor{blue}{1.00})                           & \multicolumn{1}{c|}{1.23 (\textcolor{blue}{18.07})}           & \textcolor{blue}{6.20}             \\ [2pt]
IU              & 1.94 (\textcolor{blue}{3.96})                           & 98.37 (\textcolor{blue}{1.63})                           & 92.42 (\textcolor{blue}{2.16})                           & \multicolumn{1}{c|}{6.58 (\textcolor{blue}{12.72})}           & \textcolor{blue}{5.12}             \\ [2pt]
BE               & 2.08 (\textcolor{blue}{3.82})                           & 98.07 (\textcolor{blue}{1.93})                           & 92.35 (\textcolor{blue}{2.23})                           & \multicolumn{1}{c|}{9.69 (\textcolor{blue}{9.61})}           & \textcolor{blue}{4.40}             \\ [2pt]
BS               & 0.54 (\textcolor{blue}{5.36})                           &  99.58 (\textcolor{blue}{0.42})                           & 95.33 (\textcolor{blue}{0.75})                           & \multicolumn{1}{c|}{6.52 (\textcolor{blue}{12.78})}           & \textcolor{blue}{4.83}             \\ [2pt]
L1-sparse                & 5.41 (\textcolor{blue}{0.49})                           & 99.23 (\textcolor{blue}{0.77})                           & 92.13 (\textcolor{blue}{2.45})                           & \multicolumn{1}{c|}{12.01 (\textcolor{blue}{7.29})}          & \textcolor{blue}{2.75}             \\ [2pt]
SalUn                    & 5.95 (\textcolor{blue}{0.05})                           & 96.28 (\textcolor{blue}{3.72})                           & 93.12 (\textcolor{blue}{1.46})                           & \multicolumn{1}{c|}{25.42 (\textcolor{blue}{6.12})}          & \textcolor{blue}{2.84}             \\ [2pt] \midrule [1pt]
\textbf{UFG}         & \textbf{5.56 (\textcolor{blue}{0.34})}                   & \textbf{99.92 (\textcolor{blue}{0.08})}                   & \textbf{93.14 (\textcolor{blue}{1.44})}                   & \multicolumn{1}{c|}{\textbf{8.96 (\textcolor{blue}{10.34})}} & \textbf{\textcolor{blue}{3.05}}    \\ [2pt]
\textbf{CUFG}     & \textbf{6.06 (\textcolor{blue}{0.16})}                   & \textbf{99.78 (\textcolor{blue}{0.22})}                   & \textbf{93.10 (\textcolor{blue}{1.48})}                   & \multicolumn{1}{c|}{\textbf{10.36 (\textcolor{blue}{8.94})}} & \textbf{\textcolor{blue}{2.70}}    \\ [2pt] \bottomrule [2pt]
\end{tabular}
\end{table}

\vspace{0.01cm}

\renewcommand{\thetable}{A.5}
\begin{table}[htbp]
\centering
\caption{Comparison of MU performance among various methods on the CIFAR-10 dataset using VGG-16 as the backbone network, under the forgetting scenario: Random Data Forgetting 50\%.}
\begin{tabular}{@{}p{2cm}<{\centering}|p{2cm}<{\centering}p{2cm}<{\centering}p{2cm}<{\centering}p{2cm}<{\centering}p{2cm}<{\centering}@{}}
\toprule [2pt]
\multirow{2}{*}{\textbf{Methods}} & \multicolumn{5}{c}{\textbf{Random Data Forgetting (50\%) on CIFAR-10 with VGG-16}}                                                                                                \\ [2pt]
                         & \multicolumn{1}{c|}{\textbf{UA}} & \multicolumn{1}{c|}{\textbf{RA}} & \multicolumn{1}{c|}{\textbf{TA}} & \multicolumn{1}{c|}{\textbf{MIA}}   & \textbf{Avg.Gap} \\ [2pt] \midrule [1pt]
Retrain                  & 7.86 (\textcolor{blue}{0.00})                           & 100.00 (\textcolor{blue}{0.00})                           & 91.56 (\textcolor{blue}{0.00})                           & \multicolumn{1}{p{2cm}<{\centering}|}{19.09 (\textcolor{blue}{0.00})}          & \textcolor{blue}{0.00}            \\ [2pt]
FT                 & 1.12 (\textcolor{blue}{6.74})                           & 99.72 (\textcolor{blue}{0.28})                           & 92.54 (\textcolor{blue}{0.98})                           & \multicolumn{1}{c|}{3.08 (\textcolor{blue}{16.01})}           & \textcolor{blue}{6.00}            \\ [2pt]
GA               & 1.84 (\textcolor{blue}{6.02})                           & 98.28 (\textcolor{blue}{1.72})                           & 91.87 (\textcolor{blue}{0.31})                           & \multicolumn{1}{c|}{2.53 (\textcolor{blue}{16.56})}           & \textcolor{blue}{6.15}             \\ [2pt]
IU              & 5.30 (\textcolor{blue}{2.56})                           & 94.93 (\textcolor{blue}{5.07})                           & 87.38 (\textcolor{blue}{4.18})                           & \multicolumn{1}{c|}{8.14 (\textcolor{blue}{10.95})}           & \textcolor{blue}{5.69}             \\ [2pt]
BE               & 20.58 (\textcolor{blue}{12.72})                           & 79.40 (\textcolor{blue}{20.60})                           & 72.58 (\textcolor{blue}{18.98})                           & \multicolumn{1}{c|}{11.74 (\textcolor{blue}{7.35})}           & \textcolor{blue}{14.91}             \\ [2pt]
BS               & 2.44 (\textcolor{blue}{5.42})                           &  97.58 (\textcolor{blue}{2.42})                           & 89.79 (\textcolor{blue}{1.77})                           & \multicolumn{1}{c|}{4.93 (\textcolor{blue}{14.16})}           & \textcolor{blue}{5.94}             \\ [2pt]
L1-sparse                & 4.95 (\textcolor{blue}{2.91})                           & 96.87 (\textcolor{blue}{3.13})                           & 88.61 (\textcolor{blue}{2.95})                           & \multicolumn{1}{c|}{9.39 (\textcolor{blue}{9.70})}          & \textcolor{blue}{4.67}             \\ [2pt]
SalUn                    & 3.01 (\textcolor{blue}{4.85})                           & 98.45 (\textcolor{blue}{1.55})                           & 90.37 (\textcolor{blue}{1.19})                           & \multicolumn{1}{c|}{7.77 (\textcolor{blue}{11.32})}          & \textcolor{blue}{4.72}             \\ [2pt] \midrule [1pt]
\textbf{UFG}         & \textbf{3.82 (\textcolor{blue}{4.04})}                   & \textbf{99.28 (\textcolor{blue}{0.72})}                   & \textbf{90.86 (\textcolor{blue}{0.70})}                   & \multicolumn{1}{c|}{\textbf{7.00 (\textcolor{blue}{12.09})}} & \textbf{\textcolor{blue}{4.39}}    \\ [2pt]
\textbf{CUFG}     & \textbf{4.38 (\textcolor{blue}{3.48})}                   & \textbf{98.85 (\textcolor{blue}{1.15})}                   & \textbf{90.38 (\textcolor{blue}{1.18})}                   & \multicolumn{1}{c|}{\textbf{8.27 (\textcolor{blue}{10.82})}} & \textbf{\textcolor{blue}{4.16}}    \\ [2pt] \bottomrule [2pt]
\end{tabular}
\end{table}

\vspace{0.01cm}

\renewcommand{\thetable}{A.6}
\begin{table}[htbp]
\centering
\caption{Comparison of MU performance among various methods on the CIFAR-100 dataset using ResNet-18 as the backbone network, under the forgetting scenario: Class-wise Forgetting.}
\begin{tabular}{@{}p{2cm}<{\centering}|p{2cm}<{\centering}p{2cm}<{\centering}p{2cm}<{\centering}p{2cm}<{\centering}p{2cm}<{\centering}@{}}
\toprule [2pt]
\multirow{2}{*}{\textbf{Methods}} & \multicolumn{5}{c}{\textbf{Class-wise Forgetting on CIFAR-100 with ResNet-18}}                                                                                                \\ [2pt]
                         & \multicolumn{1}{c|}{\textbf{UA}} & \multicolumn{1}{c|}{\textbf{RA}} & \multicolumn{1}{c|}{\textbf{TA}} & \multicolumn{1}{c|}{\textbf{MIA}}   & \textbf{Avg.Gap} \\ [2pt] \midrule [1pt]
Retrain                  & 100.00 (\textcolor{blue}{0.00})                           & 99.97 (\textcolor{blue}{0.00})                           & 74.90 (\textcolor{blue}{0.00})                           & \multicolumn{1}{p{2cm}<{\centering}|}{100.00 (\textcolor{blue}{0.00})}          & \textcolor{blue}{0.00}            \\ [2pt]
FT                 & 18.89 (\textcolor{blue}{81.11})                           & 99.88 (\textcolor{blue}{0.09})                           & 75.04 (\textcolor{blue}{0.14})                           & \multicolumn{1}{c|}{71.11 (\textcolor{blue}{28.89})}           & \textcolor{blue}{27.56}            \\ [2pt]
GA               & 91.56 (\textcolor{blue}{8.44})                           & 83.82 (\textcolor{blue}{16.15})                           & 60.15 (\textcolor{blue}{14.75})                           & \multicolumn{1}{c|}{96.00 (\textcolor{blue}{4.00})}           & \textcolor{blue}{10.84}             \\ [2pt]
IU              & 87.56 (\textcolor{blue}{12.44})                           & 93.21 (\textcolor{blue}{6.76})                           & 66.93 (\textcolor{blue}{7.97})                           & \multicolumn{1}{c|}{97.11 (\textcolor{blue}{2.89})}           & \textcolor{blue}{7.52}             \\ [2pt]
BE               & 60.24 (\textcolor{blue}{39.76})                           & 98.34 (\textcolor{blue}{1.63})                           & 71.96 (\textcolor{blue}{2.94})                           & \multicolumn{1}{c|}{91.88 (\textcolor{blue}{8.12})}           & \textcolor{blue}{13.11}             \\ [2pt]
BS               & 60.46 (\textcolor{blue}{39.54})                           &  98.16 (\textcolor{blue}{1.81})                           & 72.03 (\textcolor{blue}{2.87})                           & \multicolumn{1}{c|}{91.86 (\textcolor{blue}{8.14})}           & \textcolor{blue}{13.09}             \\ [2pt]
L1-sparse                & 98.67 (\textcolor{blue}{1.33})                           & 93.45 (\textcolor{blue}{6.52})                           & 68.94 (\textcolor{blue}{5.96})                           & \multicolumn{1}{c|}{100.00 (\textcolor{blue}{0.00})}          & \textcolor{blue}{3.45}             \\ [2pt]
SalUn                    & 96.89 (\textcolor{blue}{3.11})                           & 99.84 (\textcolor{blue}{0.13})                           & 75.56 (\textcolor{blue}{0.66})                           & \multicolumn{1}{c|}{100.00 (\textcolor{blue}{0.00})}          & \textcolor{blue}{0.98}             \\ [2pt] \midrule [1pt]
\textbf{UFG}         & \textbf{99.56 (\textcolor{blue}{0.44})}                   & \textbf{99.28 (\textcolor{blue}{0.69})}                   & \textbf{74.00 (\textcolor{blue}{0.90})}                   & \multicolumn{1}{c|}{\textbf{100.00 (\textcolor{blue}{0.00})}} & \textbf{\textcolor{blue}{0.51}}    \\ [2pt]
\textbf{CUFG}     & \textbf{98.89 (\textcolor{blue}{1.11})}                   & \textbf{99.81 (\textcolor{blue}{0.16})}                   & \textbf{74.71 (\textcolor{blue}{0.19})}                   & \multicolumn{1}{c|}{\textbf{100.00 (\textcolor{blue}{0.00})}} & \textbf{\textcolor{blue}{0.37}}    \\ [2pt] \bottomrule [2pt]
\end{tabular}
\end{table}

\vspace{0.22cm}

\renewcommand{\thetable}{A.7}
\begin{table}[htbp]
\centering
\caption{Comparison of MU performance among various methods on the SVHN dataset using ResNet-18 as the backbone network, under the forgetting scenario: Class-wise Forgetting.}
\begin{tabular}{@{}p{2cm}<{\centering}|p{2cm}<{\centering}p{2cm}<{\centering}p{2cm}<{\centering}p{2cm}<{\centering}p{2cm}<{\centering}@{}}
\toprule [2pt]
\multirow{2}{*}{\textbf{Methods}} & \multicolumn{5}{c}{\textbf{Class-wise Forgetting on SVHN with ResNet-18}}                                                                                                \\ [2pt]
                         & \multicolumn{1}{c|}{\textbf{UA}} & \multicolumn{1}{c|}{\textbf{RA}} & \multicolumn{1}{c|}{\textbf{TA}} & \multicolumn{1}{c|}{\textbf{MIA}}   & \textbf{Avg.Gap} \\ [2pt] \midrule [1pt]
Retrain                  & 100.00 (\textcolor{blue}{0.00})                           & 100.00 (\textcolor{blue}{0.00})                           & 96.06 (\textcolor{blue}{0.00})                           & \multicolumn{1}{p{2cm}<{\centering}|}{100.00 (\textcolor{blue}{0.00})}          & \textcolor{blue}{0.00}            \\ [2pt]
FT                 & 4.03 (\textcolor{blue}{95.97})                           & 100.00 (\textcolor{blue}{0.00})                           & 96.18 (\textcolor{blue}{0.12})                           & \multicolumn{1}{c|}{99.98 (\textcolor{blue}{0.02})}           & \textcolor{blue}{24.03}            \\ [2pt]
GA               & 93.47 (\textcolor{blue}{6.53})                           & 99.46 (\textcolor{blue}{0.54})                           & 95.24 (\textcolor{blue}{0.82})                           & \multicolumn{1}{c|}{99.64 (\textcolor{blue}{0.36})}           & \textcolor{blue}{2.06}             \\ [2pt]
IU              & 92.04 (\textcolor{blue}{7.96})                           & 99.48 (\textcolor{blue}{0.52})                           & 95.29 (\textcolor{blue}{0.77})                           & \multicolumn{1}{c|}{100.00 (\textcolor{blue}{0.00})}           & \textcolor{blue}{2.31}             \\ [2pt]
BE               & 68.92 (\textcolor{blue}{31.08})                           & 98.71 (\textcolor{blue}{1.29})                           & 94.35 (\textcolor{blue}{1.71})                           & \multicolumn{1}{c|}{92.38 (\textcolor{blue}{7.62})}           & \textcolor{blue}{10.43}             \\ [2pt]
BS               & 69.11 (\textcolor{blue}{30.89})                           &  98.67 (\textcolor{blue}{1.33})                           & 94.42 (\textcolor{blue}{1.64})                           & \multicolumn{1}{c|}{92.22 (\textcolor{blue}{7.78})}           & \textcolor{blue}{10.41}             \\ [2pt]
L1-sparse                & 100.00 (\textcolor{blue}{0.00})                           & 98.88 (\textcolor{blue}{1.12})                           & 94.12 (\textcolor{blue}{1.94})                           & \multicolumn{1}{c|}{100.00 (\textcolor{blue}{0.00})}          & \textcolor{blue}{0.77}             \\ [2pt]
SalUn                    & 98.78 (\textcolor{blue}{1.22})                           & 100.00 (\textcolor{blue}{0.00})                           & 96.10 (\textcolor{blue}{0.04})                           & \multicolumn{1}{c|}{100.00 (\textcolor{blue}{0.00})}          & \textcolor{blue}{0.32}             \\ [2pt] \midrule [1pt]
\textbf{UFG}         & \textbf{99.90 (\textcolor{blue}{0.10})}                   & \textbf{99.98 (\textcolor{blue}{0.02})}                   & \textbf{95.33 (\textcolor{blue}{0.73})}                   & \multicolumn{1}{c|}{\textbf{99.98 (\textcolor{blue}{0.02})}} & \textbf{\textcolor{blue}{0.22}}    \\ [2pt]
\textbf{CUFG}     & \textbf{99.93 (\textcolor{blue}{0.07})}                   & \textbf{99.98 (\textcolor{blue}{0.02})}                   & \textbf{95.29 (\textcolor{blue}{0.77})}                   & \multicolumn{1}{c|}{\textbf{100.00 (\textcolor{blue}{0.00})}} & \textbf{\textcolor{blue}{0.22}}    \\ [2pt] \bottomrule [2pt]
\end{tabular}
\end{table}

\vspace{0.22cm}

\renewcommand{\thetable}{A.8}
\begin{table}[htbp]
\centering
\caption{Comparison of MU performance among various methods on the CIFAR-10 dataset using VGG-16 as the backbone network, under the forgetting scenario: Class-wise Forgetting.}
\label{A.8}
\begin{tabular}{@{}p{2cm}<{\centering}|p{2cm}<{\centering}p{2cm}<{\centering}p{2cm}<{\centering}p{2cm}<{\centering}p{2cm}<{\centering}@{}}
\toprule [2pt]
\multirow{2}{*}{\textbf{Methods}} & \multicolumn{5}{c}{\textbf{Class-wise Forgetting on CIFAR-10 with VGG-16}}                                                                                                \\ [2pt]
                         & \multicolumn{1}{c|}{\textbf{UA}} & \multicolumn{1}{c|}{\textbf{RA}} & \multicolumn{1}{c|}{\textbf{TA}} & \multicolumn{1}{c|}{\textbf{MIA}}   & \textbf{Avg.Gap} \\ [2pt] \midrule [1pt]
Retrain                  & 100.00 (\textcolor{blue}{0.00})                           & 100.00 (\textcolor{blue}{0.00})                           & 93.34 (\textcolor{blue}{0.00})                           & \multicolumn{1}{p{2cm}<{\centering}|}{100.00 (\textcolor{blue}{0.00})}          & \textcolor{blue}{0.00}            \\ [2pt]
FT                 & 55.93 (\textcolor{blue}{44.07})                           & 99.50 (\textcolor{blue}{0.50})                           & 92.42 (\textcolor{blue}{0.92})                           & \multicolumn{1}{c|}{100.00 (\textcolor{blue}{0.00})}           & \textcolor{blue}{11.37}            \\ [2pt]
GA               & 81.89 (\textcolor{blue}{18.11})                           & 97.12 (\textcolor{blue}{2.88})                           & 89.37 (\textcolor{blue}{3.97})                           & \multicolumn{1}{c|}{88.76 (\textcolor{blue}{11.24})}           & \textcolor{blue}{9.05}             \\ [2pt]
IU              & 93.42 (\textcolor{blue}{6.58})                           & 94.92 (\textcolor{blue}{5.08})                           & 87.61 (\textcolor{blue}{5.73})                           & \multicolumn{1}{c|}{96.40 (\textcolor{blue}{4.60})}           & \textcolor{blue}{5.50}             \\ [2pt]
BE               & 83.65 (\textcolor{blue}{16.35})                           & 97.22 (\textcolor{blue}{2.78})                           & 90.52 (\textcolor{blue}{2.82})                           & \multicolumn{1}{c|}{94.96 (\textcolor{blue}{5.04})}           & \textcolor{blue}{6.75}             \\ [2pt]
BS               & 83.72 (\textcolor{blue}{16.28})                           &  97.16 (\textcolor{blue}{2.84})                           & 90.57 (\textcolor{blue}{2.77})                           & \multicolumn{1}{c|}{94.87 (\textcolor{blue}{5.13})}           & \textcolor{blue}{6.76}             \\ [2pt]
L1-sparse                & 100.00 (\textcolor{blue}{0.00})                           & 97.24 (\textcolor{blue}{2.76})                           & 90.61 (\textcolor{blue}{2.73})                           & \multicolumn{1}{c|}{100.00 (\textcolor{blue}{0.00})}          & \textcolor{blue}{1.37}             \\ [2pt]
SalUn                    & 100.00 (\textcolor{blue}{0.00})                           & 98.28 (\textcolor{blue}{1.72})                           & 91.79 (\textcolor{blue}{1.55})                           & \multicolumn{1}{c|}{100.00 (\textcolor{blue}{0.00})}          & \textcolor{blue}{0.82}             \\ [2pt] \midrule [1pt]
\textbf{UFG}         & \textbf{100.00 (\textcolor{blue}{0.00})}                   & \textbf{99.14 (\textcolor{blue}{0.86})}                   & \textbf{91.27 (\textcolor{blue}{2.07})}                   & \multicolumn{1}{c|}{\textbf{100.00 (\textcolor{blue}{0.00})}} & \textbf{\textcolor{blue}{0.73}}    \\ [2pt]
\textbf{CUFG}     & \textbf{100.00 (\textcolor{blue}{0.00})}                   & \textbf{99.34 (\textcolor{blue}{0.66})}                   & \textbf{91.60 (\textcolor{blue}{1.74})}                   & \multicolumn{1}{c|}{\textbf{100.00 (\textcolor{blue}{0.00})}} & \textbf{\textcolor{blue}{0.60}}    \\ [2pt] \bottomrule [2pt]
\end{tabular}
\end{table}

\section{FUTURE WORK} \label{APP:D}
This paper proposes a fine-tuning-based stable machine unlearning (MU) method guided by the Forgetting Gradient (UFG), as well as a novel paradigm of Curriculum Unlearning (CU), realized in our CUFG framework. Both approaches offer unified and generalizable solutions for a wide range of downstream tasks, enhancing the practicality and scalability of machine unlearning.

Building upon this foundation, we further reflect on the future directions along this line of research. From the perspective of efficiency, although UFG demonstrates strong stability and unlearning accuracy, its reliance on the inference and computation of forgetting gradients leads to suboptimal overall efficiency. To address this, we plan to investigate lighter and more efficient guidance mechanisms for unlearning, aiming to accelerate gradient correction and thereby improve the overall unlearning speed. On the other hand, this work presents the first systematic theoretical and empirical validation of curriculum unlearning. CUFG introduces a novel forgetting difficulty measure and a corresponding curriculum generation strategy, providing an initial yet meaningful step toward designing realistic forgetting paths. Inspired by the success of curriculum learning in deep learning, we believe curriculum unlearning holds substantial research potential and is well-positioned to emerge as a prominent subfield within the MU community. In future work, we will further explore more effective curriculum partitioning strategies, with a focus on constructing dynamic and adaptive forgetting paths to enhance the generalization and adaptability of CUFG.

%本文提出了受遗忘梯度指引的微调式稳定MU方法以及全新的课程式遗忘范式。它们为机器遗忘的广泛下游任务提供了通用的解决方案。更进一步地，我们沿着这样地路线继续思考。一方面，UFG在遗忘效率方面由于遗忘梯度地推理与计算而并未达到最优。我们将继续尝试新的遗忘信息指引方式，以更迅捷地进行梯度修正，从而加速整体的遗忘进程。另一方面，本文所提出的课程式遗忘无论是在理论机理还是实验分析方面都被验证是有效且非常符合现实世界实际情况的。CUFG对遗忘难度度量函数，以及课程生成方法进行了初步的探索。但是正如课程学习在深度学习中所展现的那样，CU还有很大的研究潜力。它完全有潜力竞争成为MU领域的一个热门分支。下一步，我们将继续探索如何更好的划分遗忘课程。
%%%%%%%%%%%%%%%%%%%%%%%%%%%%%%%%%%%%%%%%%%%%%%%%%%%%%%%%%%%%
%\fi
\end{document}